\documentclass{new_tlp}

\usepackage{microtype}

\usepackage{amsmath,amssymb}
\usepackage{times,helvet,courier}
\usepackage[pdftex]{graphicx} 
\usepackage{url}
\usepackage{listings}
\usepackage{subfig}
\usepackage{color}

\usepackage{wasysym}
\usepackage{stmaryrd}

\newcommand{\hide}[1]{}

\newcommand{\claspfolio}{\textit{claspfolio}}

\newcommand{\clasp}{\textit{clasp}}

\newcommand{\gringo}{\textit{gringo}}

\newcommand{\satzilla}{\textit{satzilla}}
\newcommand{\isac}{\textit{isac}}

\newcommand{\hydra}{\textit{hydra}}
\newcommand{\ppfolio}{\textit{ppfolio}}

\newcommand{\tnm}{\textit{tnm}}
\newcommand{\cryptominisat}{\textit{cryptominisat}}
\newcommand{\SSS}{\textit{3S}}
\newcommand{\stonesoup}{\textit{Stone Soup}}
\newcommand{\cphydra}{\textit{cphydra}}
\newcommand{\card}[1]{|#1|}

\newcommand{\tplp}[1]{\textcolor{red}{#1}}
\renewcommand{\tplp}[1]{#1}

\newcommand{\third}[1]{\textcolor{red}{#1}}
\renewcommand{\third}[1]{#1}

\newcommand{\edhh}[1]{\textcolor{blue}{#1}}
\renewcommand{\edhh}[1]{#1}

\newcommand{\sathand}{\textit{Crafted}}
\newcommand{\satrand}{\textit{Random}}
\newcommand{\satindu}{\textit{Application}}

\newcommand{\aspset}{\textit{ASP-Set}}
\newcommand{\SSSset}{\textit{3S-Set}}
\newcommand{\cspset}{\textit{CSP-Set}}
\newcommand{\qbfset}{\textit{QBF-Set}}
\newcommand{\maxsat}{\textit{MaxSAT-Set}}

\newcommand{\sbs}{\textit{single best}}
\newcommand{\uni}{\textit{uniform}}
\newcommand{\combi}{\textit{ppfolio-like}}

\newcommand{\schedule}{\textit{aspeed}} 
\newcommand{\scheduleSP}{\textit{\schedule{}-SP}}
\newcommand{\scheduleMP}{\textit{\schedule{}-4P}}
\newcommand{\aspeed}{\schedule{}}
\newcommand{\selection}{\textit{selection}}
\newcommand{\oracle}{\textit{oracle}}

\newcommand{\minO}{\textit{heu-Min}}
\newcommand{\heuO}{\textit{heu-Opt}}

\DeclareMathOperator{\argmax}{arg\,max}
\DeclareMathOperator{\argmin}{arg\,min}

\title[\schedule{}: Solver Scheduling via Answer Set Programming]
      {\schedule{}: Solver Scheduling via Answer Set Programming\footnote{Extended version of \textit{\schedule{}: ASP based Solver Scheduling} published at ICLP'12.}}

\author[H. Hoos \and R. Kaminski \and M. Lindauer \and T. Schaub]{%
		Holger Hoos \\
		Department of Computer Science\\
		University of British Columbia,\\
		Vancouver, Canada\\
		\url{hoos@cs.ubc.ca} 
		\and Roland Kaminski\\
		Institute of Informatics,\\
		University of Potsdam, Germany\\
		\url{kaminski@cs.uni-potsdam.de}
		\and Marius Lindauer\\
		Institute of Informatics,\\
		University of Potsdam, Germany\\
		\url{manju@cs.uni-potsdam.de}
		\and Torsten Schaub\\
		Institute of Informatics,\\
		University of Potsdam, Germany\\
		\url{tosten@cs.uni-potsdam.de}
}

\submitted{27 February 2013}
\revised{09 September 2013}
\accepted{27 December 2013}

\jdate{[n/a]}
\pubyear{[n/a]}
\pagerange{\pageref{firstpage}--\pageref{lastpage}}
  
\begin{document}

\maketitle

\begin{abstract}
Although Boolean Constraint Technology has made tremendous progress over the
last decade, 
\edhh{the efficacy  of state-of-the-art solvers is known to vary
considerably across different types of problem instances
and is known to depend strongly on algorithm parameters.
This problem was addressed by means of a simple, yet effective approach
using handmade, uniform and unordered
schedules of multiple solvers in \ppfolio{}, 
which showed very impressive performance in 
the 2011 SAT Competition.}
%
Inspired by this,
we take advantage of the modeling and solving capacities of Answer Set Programming (ASP) to
automatically determine more refined, that is, non-uniform and ordered solver
schedules from existing benchmarking data.
We begin by formulating the determination of such schedules as multi-criteria
optimization problems and provide corresponding ASP encodings.
The resulting encodings are easily customizable for different settings and the
computation of optimum schedules can mostly be done in the blink of an eye,
even when dealing with large runtime data sets stemming from many solvers on
hundreds to thousands of instances.
%
%
Also, \edhh{the fact that our approach can be customized easily enabled us to 
swiftly adapt it to generate parallel schedules for multi-processor machines.}
%
\end{abstract}


\begin{keywords}
Algorithm Schedules, Answer Set Programming, Portfolio-Based Solving
\end{keywords}

\section{Introduction}

Boolean Constraint Technology has made tremendous progress over the last
decade,
leading to industrial-strength solvers.
Although this advance in technology was mainly conducted in the area of
Satisfiability Testing (SAT;~\cite{SATHandbook}),
it meanwhile also led to significant boosts in neighboring areas,
like
Answer Set Programming (ASP;~\cite{baral02a}),
Pseudo-Boolean Solving \cite[Chapter~22]{SATHandbook},
and
even (multi-valued) Constraint Solving \cite{tatakiba09a}.
However, there is a prize to pay.
Modern Boolean constraint solvers are rather sensitive to the way their search
parameters are configured.
Depending on the choice of the respective configuration,
the solver's performance may vary by several orders of magnitude.
Although this is a well-known issue,
it was impressively \edhh{illustrated} once more during the 2011 SAT Competition,
where $16$ prizes were won by the portfolio-based
solver \ppfolio~\cite{roussel11a}.
The idea underlying \ppfolio\ is very simple:
it independently runs several solvers in parallel.
If only one processing unit is available, three solvers are started.
By relying on the \edhh{process scheduling mechanism of the} operating system,
each solver gets nearly the same time to solve a given instance.
We refer to this as a uniform, unordered solver schedule.
If \edhh{several} processing units are available,
\edhh{one solver is started on each unit;
however, multiple solvers may end up on the last unit.}

Inspired by this \edhh{simple, yet effective} system,
we \edhh{devise} a more elaborate, yet still simple approach that takes advantage of
the modeling and solving capacities of ASP to automatically determine more
refined, that is, non-uniform and ordered solver schedules from existing
benchmarking data.
The resulting encodings are easily customizable for different settings.
For instance, our approach is directly extensible to the generation of parallel
schedules for multi-processor machines.
Also, the computation of optimum schedules can mostly be done in the blink of an eye,
even when dealing with large runtime data sets stemming from many solvers on
hundreds to thousands of instances.
Despite its simplicity, our approach matches the performance of much
more sophisticated ones, such as \satzilla~\cite{xuhuhole08a} and \SSS~\cite{kamasasase11a}.
Unlike both, our approach does not rely on the availability of domain-specific features of the problem instance being solved,
which makes it easily adaptable to other domains.

The remainder of this article is structured as follows.
In Section~\ref{sec:seq:scheduling}, 
we formulate the problem of determining optimum schedules 
as \edhh{a multi-criteria optimization problem}.
In doing so, our primary emphasis lies in producing robust schedules that
aim at the fewest number of timeouts by non-uniformly attributing each solver
(or solver configuration) a different time slice.
Once such a robust schedule is found,
we optimize its runtime by selecting the best solver alignment.
We then extend this approach to parallel settings
in which multiple processing units are available.
With these
\edhh{formalizations} at hand,
we proceed in two steps.
First, 
we provide an ASP encoding for computing (parallel) timeout-minimal schedules 
(Section~\ref{sec:seq:asp}).
Once such a schedule is identified,
we use a second encoding to find a time-minimal
alignment of its solvers (Section \ref{sec:par:asp:time}).
Both ASP encodings are also of interest from an ASP modelling perspective,
because they reflect interesting features needed for dealing with large
sets of (runtime) data.
Finally, 
in Section~\ref{sec:experiments}, we provide an empirical evaluation of
the resulting system \schedule, and 
we contrast it with related approaches (Section~\ref{sec:related:work}).
In what follows, we presuppose a basic acquaintance with ASP
(see \cite{gekakasc12a} for a comprehensive introduction).


\section{Solver Scheduling}
\label{sec:seq:scheduling}

In the following,
we formulate the optmization problem of computing a solver schedule.
To this end,
we introduce robust timeout-minimal schedules for single-threaded sytems
that are extended by a solver alignment mechanism to minimize the used runtime.
Furthermore, in order to exploit the increasing prevalence of multi-core processors,
we consider the problem of finding good parallel solver schedules.


\subsection{Sequential Scheduling} 

Given a set $I$ of problem instances
and   a set $S$ of solvers,
we use function~$t: I \times S \mapsto \mathbb{R}\tplp{^+}$ 
to represent a table of solver runtimes on instances.
Also, we use an integer $\kappa$ to represent a given cutoff time.
For illustration, consider the runtime function in
Table~\ref{tab:seq:setting};
it deals with 6 problem instances, $i_1$ to~$i_6$, and 3 solvers,
$s_1$, $s_2$, and $s_3$.
\begin{table}
\[
\begin{array}{| r | c  c  c || c |}
  \cline{1-5}
      & s_1 & s_2 & s_3 & oracle\\
  \cline{1-5}
  i_1 & \mathbf{ 1} &         \geq 10  &          3  & 1 \\
  i_2 &          5  &         \geq 10  & \mathbf{ 2} & 2 \\
  i_3 &          8  & \mathbf{ 1} &         \geq 10  & 1 \\
  i_4 &         \geq 10  &         \geq 10  & \mathbf{ 2} & 2 \\
  i_5 &         \geq 10  & \mathbf{ 6} &         \geq 10  & 6 \\
  i_6 &         \geq 10  &          8  &         \geq 10  & 8 \\
  \cline{1-5}
  \text{timeouts} & 3 & 3 & 3 & 0 \\ 
  \cline{1-5}
\end{array}
\]
\caption{Table of solver runtimes on problem instances with $\kappa=10$; \tplp{'$\geq 10$'} indicates a timeout.}
\label{tab:seq:setting}
\end{table}

%
Each solver can solve three out of six instances within the cutoff time, $\kappa=10$;
timeouts are indicated by '$\geq 10$' in Table~\ref{tab:seq:setting}.
The oracle solver, \edhh{also known as virtual best solver (VBS)}, is obtained by assuming
the best performance of each individual solver.
As we see in the rightmost column,
the oracle would be able to solve all instances in our example within the
cutoff time;
thus, if we knew beforehand which solver to choose for each instance,
we could solve all of them.
While we can hardly hope to practically realize an oracle solver \tplp{on a single threaded system (at least in terms of CPU time)},
performance improvements can already be obtained by successively running each solver for 
a limited period of time rather than running a single solver until the cutoff is reached.
For instance, by \tplp{uniformly distributing} time over all three solvers in our example, as done in \ppfolio{},
we could solve $4$ out of $6$ instances, namely instance $i_1 \ldots i_4$.
Furthermore, the number of solved instances can be increased further by 
running $s_1$ for $1$, $s_2$ for $6$, and $s_3$ for $2$ seconds,
which allows us to solve $5$ out of $6$ instances, 
as indicated in bold in Table~\ref{tab:seq:setting}.
In what follows,
we show how such a optimized non-uniform schedule can be obtained beforehand from given runtime data.

Given $I$,  $S$, $t$, and $\kappa$ as specified above,
a \emph{timeout-optimal solver schedule} can be expressed as
a function $\sigma:S\rightarrow[0,\kappa]$,
satisfying the following condition:

\begin{equation}
  \label{eq:criterion:one}
  \begin{array}{c}
    \sigma
    \in
    \argmax_{\sigma:S\rightarrow[0,\kappa]}
    \card{\{i\mid \exists s\in S: t(i,s)\leq\sigma(s)\}}
    \\[5pt]
    \text{ such that }\qquad
    \textstyle{\sum}_{s\in S}{\sigma(s)} \leq \kappa
  \end{array}
\end{equation}

An optimal schedule $\sigma$ consists of slices $\sigma(s)$ indicating the
(possibly zero) time allotted to each solver $s\in S$.
Such a schedule maximizes the number of solved instances, or conversely,
minimizes the number of timeouts.
\tplp{An instance $i$ is solved by $\sigma$ 
if there is a solver $s \in S$ 
that has an equal or greater time slice $\sigma(s)$
than the time needed by the solver to solve the instance, viz. $t(i,s)$.
As a side constraint, 
the sum of all time slices $\sigma(s)$ has to be equal or less than the cutoff time $\kappa$.} 

The above example corresponds to the schedule $\sigma=\{s_1 \mapsto 1,s_2 \mapsto 6,s_3 \mapsto 2\}$;
in fact, $\sigma$ constitutes one of nine timeout-optimal solver schedules
in our example.
Note that the sum of all time slices is even smaller than the
cutoff time.
Hence, all schedules obtained by adding 1 to either of the three solvers are also timeout-optimal.
A timeout-optimal schedule consuming the entire allotted time is $\{s_1 \mapsto 0,s_2 \mapsto 8,s_3 \mapsto 2\}$.

In practice, however, the criterion in \eqref{eq:criterion:one} turns out to be
too coarse, that is, it \edhh{often admits a diverse set of} solutions among which we would
like to make an educated choice.%
\hide{
\footnote{For instance, on the benchmark classes used in
  Section~\ref{sec:experiments}, \textit{clasp} finds 
  331.642 (\satrand), 220.693 (\sathand), and 253.124 (\satindu) 
  timeout-optimal schedules within one minute, respectively.}
}
To this end,
we \edhh{make use of (simplified) $L$-norms as the basis for refining our choice of schedule.}
In our case,
an $L^n$-norm on schedules is defined%
\footnote{The common $L^n$-norm is defined as $\sqrt[n]{\Sigma_{x\in X}{x}^n}$.
We take the simpler definition in view of using it merely for optimization.}
as
\(
\Sigma_{s\in S,\sigma(s)\neq 0\,}{\sigma(s)}^n
\).
Depending on the choice of $n$ as well as whether we minimize or maximize the
norm,
we obtain different selection criteria.
For instance, 
$L^0$-norms suggest using as few (or as many) solvers as possible,
and
$L^1$-norms aim at minimizing (or maximizing) the sum of time slices.
Minimizing the $L^2$-norm amounts to allotting each solver a similar time
slice,
while maximizing it prefers schedules with large runtimes for few solvers.
\edhh{In more formal terms, 
for a given set $S$ of solvers,
using an $L^n$-norm we would like to determine schedules satisfying the constraint}
\begin{equation}
  \label{eq:norms}
  \sigma\in\argmin_{\sigma:S\rightarrow[0,\kappa]}
  \Sigma_{s\in S,\sigma(s)\neq 0\,}{\sigma(s)}^n,
\end{equation}
\edhh{or the analogous constraint for $\argmax$ (in case of maximization)}.

For instance,
our example schedule
$\sigma=\{s_1 \mapsto 1,s_2 \mapsto 6,s_3 \mapsto 2\}$ has the $L^n$-norms
3, 9, and 41 for $n=0..2$.
In contrast,
we obtain norms 3, 9, and 27
for the (suboptimal) uniform schedule $\{s_1 \mapsto 3,s_2 \mapsto 3,s_3 \mapsto 3\}$
and 1, 9, and 81
for a singular schedule $\{s_3 \mapsto 9\}$, 
respectively.
Although empirically, we found that schedules for various $n$
as well as for minimization and maximization have useful properties,
overall, we favor schedules with a minimal $L^2$-norm.
First, this choice leads to a significant reduction of candidate schedules
and,
second, it results in schedules with a maximally homogeneous distribution of time slices, 
similar to \ppfolio.
In fact, our example schedule has the smallest $L^2$-norm
among all nine timeout-optimal solver schedules.

Once we have identified \edhh{an optimal} schedule \edhh{w.r.t.{}} criteria
\eqref{eq:criterion:one} and  \eqref{eq:norms},
it is interesting to determine which solver alignment yields the best
performance as regards time.
More formally,
we define an \emph{alignment} of a set $S$ of solvers as a bijective function
\(
\pi: \{1,\dots,|S|\}\rightarrow S
\).
Consider the above schedule $\sigma=\{s_1 \mapsto 1,s_2 \mapsto 6,s_3 \mapsto 2\}$.
The alignment $\pi=\{1\mapsto s_1, 2\mapsto s_3, 3\mapsto s_2\}$ 
induces the execution sequence $(s_1, s_3, s_2)$ of $\sigma$.
This sequence takes $29$ seconds for all six benchmarks in Table~\ref{tab:seq:setting}; 
in detail, it takes $1, 1+2, 1+2+1, 1+2, 1+2+6, 1+2\tplp{+7}$ seconds for
benchmark $i_k$ for $k=1..6$,
whereby instance $i_6$ could not be solved.
For instance, benchmark $i_3$ is successfully solved by the third solver in the
alignment, viz.~$s_2$.
Hence the total time amounts to the time allotted by $\sigma$ to $s_1$ and $s_3$,
viz.~$\sigma(s_1)$ and $\sigma(s_3)$,
plus the effective time of $s_2$, viz.~$t(i_3,s_2)$.

This can be \edhh{formalized} as follows.
Given a schedule $\sigma$\ and an alignment $\pi$ of a set $S$ of solvers, 
and an instance $i\in I$,
we define the runtime $\tau$ of schedule $\sigma$ aligned by $\pi$ on $i$:
\begin{equation}
  \label{eq:criterion:two:aux}
  \tau_{\sigma,\pi}(i) = 
    \begin{cases} 
      \left(
        \textstyle{\sum}_{j=1}^{\min{(P_{\sigma,\pi})} - 1}
        \sigma(\pi(j))
      \right)
      +
      t(i,\pi(\min{(P_{\sigma,\pi})}))
       & \text{if $P_{\sigma,\pi} \not= \emptyset$,}\\
      \kappa & \text{otherwise}
    \end{cases}
\end{equation}
where
\(
P_{\sigma,\pi} = \{ l \in \{1,\dots,|S|\} \mid t(i,\pi(l))\leq\sigma(\pi(l)) \}
\) are the positions of solvers solving instance $i$ in a schedule $\sigma$ aligned by $\pi$.
%
If an instance $i$ cannot be solved at all by a schedule,
$\tau_{\sigma,\pi}(i)$ is set to the cutoff $\kappa$.
For our example schedule $\sigma$ and its alignment $\pi$,
we obtain for $i_3$: $\min P_{\sigma,\pi}=3$ and $\tau_{\sigma,\pi}(i_3)=1+2+1=4$.

For a schedule $\sigma$ of solvers in $S$, we then define the optimal aligment of schedule $\sigma$:
\begin{equation}
  \label{eq:criterion:two}
  \pi
  \in
  \argmin_{\pi:\{1,\dots,|S|\}\rightarrow S}
  \textstyle{\sum}_{i\in I}
  \tau_{\sigma,\pi}(i)
\end{equation}

For our timeout-optimal schedule $\sigma=\{s_1~\mapsto~1, s_2~\mapsto~6, s_3~\mapsto~2\}$
w.r.t.{} criteria~\eqref{eq:criterion:one} and  \eqref{eq:norms},
we obtain two optimal execution alignments,
namely $(s_3,s_1,s_2)$ and $(s_1,s_3,s_2)$, both of which result in a solving time of $29$ seconds for the benchmarks of Table~\ref{tab:seq:setting}.


\subsection{Parallel Scheduling}\label{sec:par:scheduling}
The increasing availability of multi-core processors makes it interesting
to extend our approach for distributing a schedule's solvers \edhh{over multiple}
processing units.
For simplicity, we take a coarse approach in binding solvers to units,
thus precluding re-allocations during runtime.

To begin with,
let us provide a formal specification of the extended problem.
To this end,
we augment our \edhh{previous formalization} with a set $U$ of (processing) units
and associate each unit with subsets of solvers from $S$.
More formally,
we define a \emph{distribution} of a set $S$ of solvers as the function
\(
\eta: U\rightarrow 2^S
\) 
such that $\bigcap_{u \in U}\eta(u) = \emptyset$.
With it,
we can determine timeout-optimal solver schedules for several cores simply by
strengthening the condition in~\eqref{eq:criterion:one}  to the effect that
all solvers associated with the same unit must respect the cutoff time.
This leads us to the following extension of~\eqref{eq:criterion:one}:
\begin{equation}
  \label{eq:criterion:one:parallel}
  \begin{array}{c}
    \sigma
    \in
    \argmax_{\sigma:S\rightarrow[0,\kappa]}
    \card{\{i\mid \exists s\in S: t(i,s)\leq\sigma(s)\}}
    \\[5pt]
    \text{ such that }\qquad
    \textstyle{\sum}_{s\in\eta(u)}{\sigma(s)} \leq \kappa
    \ \text{ for each }u\in U    
  \end{array}
\end{equation}

For illustration,
let us reconsider Table~\ref{tab:seq:setting} along with schedule $\sigma=\{s_1 \mapsto 1, s_2 \mapsto 8, s_3 \mapsto 2\}$.
Assume that we have two cores, 1 and 2, along with the distribution
\(
\eta=\{1\mapsto\{s_2\}, 2\mapsto\{s_1,s_3\}\}
\).
This distributed schedule is an optimal solution to the optimization problem in~\eqref{eq:criterion:one:parallel} 
w.r.t.{} the benchmarks in Table~\ref{tab:seq:setting}
because it solves all benchmarks within a cutoff time of $\kappa~=~8$.

We keep the definitions of a schedule's $L^n$-norm as a global constraint.
However, for determining our secondary criterion, enforcing time-optimal schedules,
we relativize the auxiliary definitions in \eqref{eq:criterion:two:aux}
to account for each unit separately.
Given a schedule $\sigma$ and a set $U$ of processing units,
we define for each unit $u\in U$ a \emph{local alignment} of the solvers in
$\eta(u)$ as the bijective function
\(
\pi_u: \{1,\dots,|\eta(u)|\}\rightarrow\eta(u)
\).
\edhh{Given this function and a problem instance} $i\in I$,
we extend the definitions in \eqref{eq:criterion:two:aux} as follows:
\begin{equation}
  \label{eq:criterion:two:parallel:aux}
  \tau_{\sigma,\pi_u}(i) = 
    \begin{cases} 
      \left(
        \textstyle{\sum}_{j=1}^{\min{(P_{\sigma,\pi})} - 1}
        \sigma(\pi_u(j))
      \right)
      +
      t(i,\pi_u(\min{(P_{\sigma,\pi})}))
       & \text{if $P_{\sigma,\pi} \not= \emptyset$,}\\
      \kappa & \text{otherwise}
    \end{cases}
\end{equation}
where $P_{\sigma,\pi} = \{ l \in \{1,\dots,|\eta(u)|\} \mid
  t(i,\pi_u(l))\leq\sigma(\pi_u(l)) \}$.

The collection
\(
(\pi_u)_{u\in U}
\)
regroups all local alignments into a \emph{global alignment}. 
For a schedule $\sigma$ of solvers in $S$ and a set $U$ of (processing) units,
we then define an optimal global alignment:
\begin{equation}
  \label{eq:criterion:two:parallel}
  (\pi_u)_{u\in U}
  \in
  \argmin_{(\pi_u: \{1,\dots,|\eta(u)|\}\rightarrow\eta(u))_{u\in U}}
  \textstyle{\sum}_{i\in I}
  \min_{u\in U}\tau_{\sigma,\pi_u}(i)
\end{equation}
For illustration,
reconsider the above schedule $\sigma=\{s_1 \mapsto 1, s_2 \mapsto 8, s_3 \mapsto 2\}$
and distribution \( \eta=\{1\mapsto\{s_2\}, 2\mapsto\{s_1,s_3\}\}\),
and suppose we chose the local alignments
\tplp{
\(
\pi_1=\{1 \mapsto s_2\}
\)
and
\(
\pi_2=\{1 \mapsto s_1, 2\mapsto s_3\}
\).}
This global alignment solves all six benchmark instances of Table~\ref{tab:seq:setting} in 22 seconds wallclock time.
In more detail,
it takes $1_2, 1+2_2, 1_1, 1+2_2, 6_1, 8_1$ seconds for
instance $i_k$ for $k=1..6$,
where the solving unit is indicated by the subscript.

Note that the definitions in 
\eqref{eq:criterion:one:parallel},
\eqref{eq:criterion:two:parallel:aux},
and
\eqref{eq:criterion:two:parallel}
correspond to their sequential counterparts
in
\eqref{eq:criterion:one},
\eqref{eq:criterion:two:aux},
and
\eqref{eq:criterion:two}
whenever we are faced with a single processing unit.


\section{Solving Timeout-Optimal Scheduling with ASP}
\label{sec:seq:asp}

To begin with,
we detail the basic encoding for identifying robust (parallel) schedules.
In view of the remark at the end of the last section,
however,
we directly provide an encoding for parallel scheduling,
which collapses to one for sequential scheduling 
whenever a single processing unit is used.

Following good practice in ASP,
a problem instance is expressed as a set of facts.
That is,
Function $t: I \times S \mapsto \mathbb{R}$ is represented as facts of form \texttt{time(i,s,t)},
where $\texttt{i}\in I$, $\texttt{s}\in S$, and $\texttt{t}$ is the runtime
$\texttt{t}(\texttt{i},\texttt{s})$, converted to a natural number with limited precision.
The cutoff is expressed via Predicate~\texttt{kappa/1},
and the number of available processing units is captured via Predicate~\texttt{units/1},
here instantiated to $2$ units.
Given this, we can represent the contents of Table~\ref{tab:seq:setting} as
shown in Listing~\ref{lst:facts} below.
\begin{lstlisting}[numbers=none,caption=Facts, label=lst:facts]
kappa(10).
units(2).

time(i1, s1,  1).  time(i1, s2, 11).  time(i1, s3,  3).
time(i2, s1,  5).  time(i2, s2, 11).  time(i2, s3,  2).
time(i3, s1,  8).  time(i3, s2,  1).  time(i3, s3, 11).
time(i4, s1, 11).  time(i4, s2, 11).  time(i4, s3,  2).
time(i5, s1, 11).  time(i5, s2,  6).  time(i5, s3, 11).
time(i6, s1, 11).  time(i6, s2,  8).  time(i6, s3, 11).
\end{lstlisting}

The encoding in Listing~\ref{lst:enc} along with all following ones are given in
the input language of \textit{gringo}~\cite{potasscoManual}.
The first three lines of Listing~\ref{lst:enc} provide auxiliary data.
The set $S$ of solvers is given by Predicate \texttt{solver/1}.
Similarly, the runtimes for each solver are expressed by \texttt{time/2}
and each processing unit by \texttt{unit/1}.
In addition, the ordering of instances by time per solver is precomputed;
it is expressed via \texttt{order/3}.
\begin{lstlisting}[numbers=none,caption=\texttt{I} is solved immediatly before \texttt{J} by solver \texttt{S}]
order(I,J,S) :-
  time(I,S,T), time(J,S,V), (T,I) < (V,J),
  not time(K,S,U) : time(K,S,U) : (T,I) < (U,K) : (U,K) < (V,J).
\end{lstlisting}
The above results in facts \texttt{order(I,J,S)} 
capturing that 
instance \texttt{J} follows instance \texttt{I} by sorting the instances according to their runtimes.
Although this information could be computed via ASP (as shown above),
we make use of external means for sorting 
(the above rule needs cubic time for instantiation, 
which is infeasible for a few thousand instances).
Instead, we use \textit{gringo}'s embedded scripting language \textit{lua} for sorting.

The idea of Listing~\ref{lst:enc} is now to guess for each solver a time slice and a processing unit (in Line~$5$).
With the resulting schedule,
all solvable instances can be identified (in Line~$10-12$),
and finally, all schedules solving \edhh{a maximal number of} instances are selected (in Line~$14$).

\begin{lstlisting}[%
caption=ASP encoding for Timeout-Minimal (Parallel) Scheduling,
label=lst:enc%
]
solver(S)  :- time(_,S,_).
time(S,T)  :- time(_,S,T).
unit(1..N) :- units(N).

{slice(U,S,T): time(S,T): T <= K: unit(U)} 1 :- solver(S),kappa(K).
 :- not [ slice(U,S,T) = T ] K, kappa(K), unit(U).

slice(S,T) :- slice(_,S,T).

solved(I,S) :- slice(S,T), time(I,S,T).
solved(I,S) :- solved(J,S), order(I,J,S).
solved(I)   :- solved(I,_).

#maximize { solved(I) @ 2 }.  
#minimize [ slice(S,T) = T*T @ 1 ].
\end{lstlisting}%
In more detail, a schedule is represented by atoms \texttt{slice(U,S,T)}
allotting a time slice \texttt{T} to solver \texttt{S} on unit \texttt{U}.
In Line~$5$,
at most one time slice is chosen for each solver,
subject to the condition that it \edhh{does not exceed} the cutoff time.
At the same time, a processing unit is uniquely assigned to the selected solver.
The integrity constraint in Line~6 ensures that the sum over all selected time
slices on each processing unit is not greater than the cutoff time.
This implements the side condition in~\eqref{eq:criterion:one:parallel},
and it reduces to the one in~\eqref{eq:criterion:one} whenever a single unit is considered.
The next  line projects out the processing unit 
because it is irrelevant when determining solved instances (in Line~8).
In Lines~10 to 12, 
all instances solved by the selected time slices are gathered via predicate \texttt{solved/1}. 
\edhh{Considering} that we collect in Line~8 all time slices among actual runtimes,
each time slice allows for solving at least one instance.
This property is used in Line~10 to identify the instance \texttt{I} solvable by solver \texttt{S};
\edhh{using it, along with} the sorting of instances by solver performance in \texttt{order/3},
we collect in Line~11 all instances that can be solved even faster than the instance in Line~10.
Note that at first sight it might be tempting to encode Lines~$10--12$ differently:
\begin{lstlisting}[numbers=none]
solved(I) :- slice(S,T), time(I,S,TS), T <= TS.
\end{lstlisting}
The problem with the above rule is that it has a quadratic number of instantiations
in the number of benchmark instances in the worst case.
\edhh{In contrast}, our ordering-based encoding is linear, because only
successive instances are considered.
Finally, the number of solved instances is maximized in Line~14,
\edhh{using the conditions from} \eqref{eq:criterion:one:parallel} 
(or \eqref{eq:criterion:one}, respectively).
This \edhh{primary objective is assigned a higher priority
than the $L^2$-norm from~\eqref{eq:norms} (priority 2 \emph{vs} 1)}.


\section{Solving (Timeout and) Time-Minimal Parallel Scheduling with ASP}
\label{sec:par:asp:time}

In the previous section,
we have \edhh{explained how to determine} a timeout-minimal (parallel) schedule.
Here, we present an encoding that takes such a schedule 
and calculates a solver alignment per processing unit
while minimizing the overall runtime 
according to Criterion~(\ref{eq:criterion:two:parallel}).
This two-phase approach is motivated by the fact that an optimal alignment must
be determined among all permutations of a schedule.
While a one-shot approach had to account for all permutations of all potential
timeout-minimal schedules,
our two-phase approach reduces the second phase to searching among all
permutations of a single timeout-minimal schedule.
\hide{
We empirically underpin this drastic reduction in Section~\ref{sec:experiments}.
}

We begin by extending the \edhh{ASP formulation}
from the last section
(in terms of \texttt{kappa/1}, \texttt{units/1}, and \texttt{time/3})
by facts over \texttt{slice/3} 
providing the time slices of a timeout-minimal schedule (per solver and processing unit).
\edhh{In the case of} our example from Section~\ref{sec:par:scheduling},
we extend the facts of Listing~\ref{lst:facts} with the following obtained timeout-minimal schedule 
to create the problem instance:
\begin{lstlisting}[numbers=none,caption=Schedule Facts, label=lst:sfacts]
slice(1,s2,8). slice(2,s1,1). slice(2,s3,2).
\end{lstlisting}

The idea of the encoding in Listing~\ref{lst:time} is to guess a permutation of solvers
and then to use ASP's optimization capacities for calculating a 
time-minimal alignment.
The challenging part is to keep the encoding compact.
That is,
we have to keep the size of the instantiation of the encoding small,
because otherwise, we \edhh{cannot hope to effectively deal with rather common situations involving} thousands of benchmark
instances.
To \edhh{this end}, we make use of \texttt{\#sum} aggregates with negative weights (Line~$23$) to
find the fastest processing unit without representing any sum of times explicitly.
\begin{figure} 
\begin{lstlisting}[%
caption=ASP encoding for Time-Minimal (Parallel) Scheduling,
label=lst:time%
]
solver(U,S)     :- slice(U,S,_).
instance(I)     :- time(I,_,_).
unit(1..N)      :- units(N).
solvers(U,N)    :- unit(U), N := {solver(U,_)}.
solved(U,S,I)   :- time(I,S,T), slice(U,S,TS), T <= TS.
solved(U,I)     :- solved(U,_,I).
capped(U,I,S,T) :- time(I,S,T), solved(U,S,I).
capped(U,I,S,T) :- slice(U,S,T), solved(U,I), not solved(U,S,I).
capped(U,I,d,K) :- unit(U), kappa(K), instance(I), not solved(U,I).
capped(I,S,T)   :- capped(_,I,S,T).

1 { order(U,S,X) : solver(U,S) } 1 :- solvers(U,N), X = 1..N.
1 { order(U,S,X) : solvers(U,N) : X = 1..N } 1 :- solver(U,S).

solvedAt(U,I,X+1) :- solved(U,S,I), order(U,S,X).
solvedAt(U,I,X+1) :- solvedAt(U,I,X), solvers(U,N), X <= N.

mark(U,I,d,K) :- capped(U,I,d,K).
mark(U,I,S,T) :- capped(U,I,S,T), order(U,S,X), not solvedAt(U,I,X).
min(1,I,S,T)  :- mark(1,I,S,T).

less(U,I) :- unit(U), unit(U+1), instance(I),
  [min(U,I,S1,T1): capped(I,S1,T1) = T1, mark(U+1,I,S2,T2) = -T2] 0.

min(U+1,I,S,T) :- min(U,I,S,T), less(U,I).
min(U,I,S,T)   :- mark(U,I,S,T), not less(U-1,I).

#minimize [min(U,_,_,T): not unit(U+1) = T].
\end{lstlisting}%
\end{figure}

The block in Line~1 to~10 gathers static knowledge about the problem instance, 
that is,
solvers per processing unit (\texttt{solver/2}),
instances appearing in the problem description (\texttt{instance/1}),
available processing units (\texttt{unit/1}),
number of solvers per unit (\texttt{solvers/2}),
instances solved by a solver within its allotted slice (\texttt{solved/3}),
and instances that could be solved on a unit given the schedule (\texttt{solved/2}).
Note that, in contrast to the previous encoding (Listing~\ref{lst:enc}),
the solved instances (\texttt{solved/3}) can be efficiently expressed as done in Line~$5$ of Listing~\ref{lst:time},
because \texttt{slice/3} are facts here.
In view of Equation~(\ref{eq:criterion:two:parallel:aux}), we precompute the times that contribute to the values of $\tau_{\sigma,\pi_u}$
and capture them in \texttt{capped/4} (and \texttt{capped/3}).
A fact \texttt{capped(U,I,S,T)} assigns to instance \texttt{I} run by solver \texttt{S} on unit \texttt{U}
a time \texttt{T}.
In Line~7, we assign the time needed to solve the instance 
if it is within the solver's time slice.
In Line~8, we assign the solver's time slice
if the instance could not be solved, but at least one other solver could solve it \edhh{on processing unit \texttt{U}}.
In Line~9, we assign the \edhh{entire} cutoff to dummy solver \texttt{d}
(we assume that there is no other solver called \texttt{d})
if the instance could not be solved on the processing unit at all;
this is to implement the else case in~(\ref{eq:criterion:two:parallel:aux}) and (\ref{eq:criterion:two:aux}).

The actual encoding starts in Line~12 and~13 by guessing a permutation of solvers.
Here, the two head aggregates ensure that for every solver (per unit) there is exactly 
one position in the alignment
and vice versa.
In Line~15 and~16, we mark indexes (per unit) as solved
if the solver with the preceding index could solve the instance
or if the previous index was marked as solved.
Note that this is a similar ``chain construction'' \edhh{used} in the previous
section in order to avoid a combinatorial blow-up.

In the block from Line~18 to~26,
we determine the time for the fastest processing unit depending on the guessed permutation.
The rules in Line~18 and~19 mark the times that have to be added up on each processing unit;
the sums of these times correspond to $\tau_{\sigma,\pi_u}(i)$ in Equation~(\ref{eq:criterion:two:parallel:aux}) and~(\ref{eq:criterion:two:aux}).
Next, we determine the smallest sum of times
\edhh{by iteratively determining the minimum}.
An atom \texttt{min(U,I,S,T)} marks the times of the fastest unit in the range from unit \texttt{1} to \texttt{U}
to solve an instance 
(or the cutoff via dummy solver \texttt{d},
if the schedule does not solve the instance for the unit).
To this end, we initialize \texttt{min/4} with the times for the first unit in Line~20.
Then, we add a rule in Line~22 and~23 that,
given minimal times for units in the range of \texttt{1} to \texttt{U} and times for unit \texttt{U+1},
determines the faster one.
The current minimum contributes positive times to the sum, 
while unit \texttt{U+1} contributes negative times.
Hence, if the sum is negative or zero,
the sum of times captured in $\texttt{min}/4$ is \edhh{smaller than or equal to} the sum of times of unit \texttt{U+1},
\edhh{and therefore,}
the unit thus slower than some preceding unit,
which makes the aggregate true and derives the corresponding atom over \texttt{less/2}.
Depending on \texttt{less/2}, we propagate the smaller sum,
which is either contributed by unit \texttt{U+1} (Line~25)
or the preceding units (Line~26).
Finally, in Line~28, the times of the fastest processing unit are minimized in the optimization statement,
which implements Equation~(\ref{eq:criterion:two:parallel}) and~(\ref{eq:criterion:two}).

\hide{
We note that the encodings given in Listing~\ref{lst:enc} and~\ref{lst:time}
are slightly simplified to highlight the crucial aspects of the fine-tuned ones
available at~\cite{aspeed}.
}

\section{Experiments}
\label{sec:experiments}

After describing the theoretical foundations and ASP encodings underlying our approach,
we now present the results from an empirical evaluation on representative ASP, \tplp{CSP, MaxSAT, SAT and QBF} benchmarks.
The python implementation of our approach, dubbed \schedule, uses 
the state-of-the-art ASP systems~\cite{contest11a} of the potassco group \cite{gekakaosscsc11a},
namely the grounder \gringo\ (3.0.4) and the ASP solver \clasp\ (2.0.5).
The sets of runtime data 
used in this work 
are freely available online.~\footnote{\url{http://www.cs.uni-potsdam.de/aspeed}}

\subsection{Experimental setup}

\begin{table}
\begin{tabular}{lcccc}
\hline\hline
   & \satrand & \sathand & \satindu & \aspset \\
 \cline{2-5}
 Cutoff (sec.) & $5000$ & $5000$ & $5000$ & $900$  \\
 \#Instances  & $600$ & $300$ & $300$  & $2589$ \\
 \#Solvers  & $9$ & $15$ & $18$ & $25$  \\
 Source & ($1$) & ($1$) & ($1$) & ($2$) \\
  \multicolumn{5}{c}{ \vspace*{-0.3cm}}\\
\hline\hline
  & \SSSset & \cspset & \qbfset & \maxsat \\
\cline{2-5}
 Cutoff (sec.) & $5000$ & $5000$ & $3600$ & $1800$\\
 \#Instances & $5467$ & $2024$ & $1368$ & $337$\\
 \#Solvers & $37$ & $2$ & $5$ & $11$\\
 Source & ($3$) & ($4$) & ($5$) & ($6$) \\
\hline\hline
\end{tabular}
\caption{Runtime data sets used in our experiments from the 2011 SAT Competition~(1), 
the ASP benchmark repository \textit{asparagus}~(2), 
\tplp{Kadioglu et~al.~2011~(3),
Gent et al. 2010~(4),
Pulina and Tacchella 2009~(5)
and Malitsky et al. 2013~(6).
} 
}
\label{tab:resources}
\end{table}

Our experiments are based on a set of runtime data 
obtained by running several solvers (or solver configurations)
on a set of benchmark instances (similar to Table~\ref{tab:seq:setting}).
To provide a thorough empirical evaluation of our approach, 
we selected eight large data sets of runtimes for \tplp{five} prominent and widely studied problems,
\tplp{ASP, CSP, MaxSAT, SAT and QBF}; these are summarized in Table~\ref{tab:resources}.
The sets \satrand, \sathand\ and \satindu\ contain the authentic runtimes 
taken from the $2011$ SAT Competition\footnote{\texttt{http://www.cril.univ-artois.fr/SAT11}}
with a cutoff of $5000$ seconds.
We selected all non-portfolio, non-parallel solvers from the main phase of the competition,
in order to provide a fair comparison with the portfolio-based SAT Solver \satzilla~\cite{xuhuhole08a},
which has been evaluated based on the same data \cite{xuhuhole11b}.

Also, we evaluated our approach on an ASP instance set (\aspset) 
based on different configurations 
of the highly parametric ASP solver \clasp~\cite{gekasc09c},
which is known to show excellent performance on a wide range of ASP instances.
We used the complementary configuration portfolio of \claspfolio{}~($1.0.1$)~\cite{gekakascsczi11a} 
designed by the main developer of \clasp{}, B.~Kaufmann, and measured the runtime of \clasp{}~($2.1.0$).
Because the instance sets from recent ASP competitions are very unbalanced~\cite{hokascsc13a}
(most of them are either too easy or too hard for \clasp{}),
we select instances from the ASP benchmark repository \textit{Asparagus},\footnote{\texttt{http://asparagus.cs.uni-potsdam.de}}
including the $2007$ (SLparse track), $2009$ and $2011$ ASP Competitions.
\gringo{} was not able to ground some instance from the $2011$ ASP Competition within $600$ CPU seconds and $2$~GB RAM,
and thus those instances were excluded.
Our \aspset{} is comprised of the $2589$ remaining instances.

\tplp{The runtime measurements for our \aspset{} were performed
on a compute cluster with $28$ nodes,
each equipped with two Intel Xeon E$5520$ $2.26$GHz quad-core CPUs and $48$ GB RAM, 
running Scientific Linux (2.6.18-308.4.1.el5).
Since all \clasp{} configurations used in our experiments are deterministic,
their runtimes on all instances were measured only once.
}

\tplp{Furthermore, we evaluated our approach on sets already used in the literature.
The set of runtime data provided by 
Kadioglu~et~al. was part of the submission of their solver \SSS~\cite{kamasasase11a}
to the $2011$ SAT Competition.
We selected this set, which we refer to as \SSSset, because it includes runtimes of many recent SAT solvers 
on prominent SAT benchmark instances.
The \cspset{} was used by~\citeN{gejekomimonipe10a},
the \qbfset{} by~\citeN{pultac09a},
and \maxsat{} by~\citeN{mamesu13a}, respectively. 
}

The performance of \schedule\ was determined from the schedules computed for Encodings \ref{lst:enc} 
and \ref{lst:time} with a minimization of the $L^2$-norm as second optimization criterion.
Although we empirically observed no clear performance gain from the latter,
we favour a schedule with a minimal $L^2$-norm:
First, it leads to a significant reduction of candidate schedules
and second, it results in schedules with a more uniform distribution of time slices, 
(resembling those used in \ppfolio).
All runtimes for the schedule computation were measured in CPU time rounded up to the next integer value, 
and runtime not allocated in the computed schedule was uniformly distributed among all solvers in the schedule.

Using the previously described data sets, we compared \schedule\ against 

\begin{itemize}
  \item \sbs: the best solver in the respective portfolio, 
  \item \uni: a uniform distribution of the time slices over all solvers in the portfolio, 
  \item \combi: an approach inspired by \ppfolio{},
  		where the best three complementary solvers are selected with an uniform distribution of time slices in the sequential case,
  \item \satzilla{}~\cite{xuhuhole11b} and \claspfolio{}~\cite{gekakascsczi11a},
  		prominent examples of model-based algorithm selection solvers for SAT and ASP, respectively,
  \item as well as against the \oracle\ performance (also called virtual best solver)\footnote{%
The performance of the \oracle\ is the minimal runtime of each instance given a portfolio of solvers and
corresponds to a portfolio-based solver with a perfect selection of the best solver for a given instance.
}.
\end{itemize}

The performance of \satzilla\ for \satrand{}, \sathand{} and \satindu{} was extracted from results reported in the literature~\cite{xuhuhole11b},
which were obtained using $10$-fold cross validation.
In the same way, \claspfolio{} was trained and cross-validated on the \aspset{}. 
In the following, 
the \selection{} approach represents \satzilla{} for the three SAT competition sets and \claspfolio{} for the \aspset.

%
Unfortunately, \schedule\ could not be directly compared against \SSS{},
because the tool used by \SSS\ to compute the underlying model is 
not freely available and hence, we were unable to train \SSS\ on new data sets.
To perform a fair comparison between \schedule\ and \SSS,
we compare both systems in an additional experiment in the last part of this section. 

\subsection{Schedule Computation}
\begin{table}
\begin{tabular}{clp{0.6em}cccc}
  \hline\hline
  \#cores & Opt. Step & &\satrand & \sathand & \satindu & \aspset\\
  \cline{4-7}
  $1$ & Schedule (sec) & & $0.54$ & $0.45$ & $119.2$ & $>1d$\\
  $1$ & Alignment (sec) & & $0.04$ & $0.23$ & $0.07$ & $0.50$\\
  \cline{4-7}
  $8$ & Schedule (sec) & & $0.28$ & $0.05$ & $61.65$ & $>1d$\\
  $8$ & Alignment (sec) & & $0.02$ & $0.006$ & $0.07$  & $0.50$\\
  \cline{4-7}
  $1$ & Combined (sec) & & $>1d$ & $47175$ & $>1d$ & $MEMOUT$\\
  \hline\hline
 					& & & \SSSset & \cspset & \qbfset & \maxsat\\
  \cline{4-7}
  $1$ & Schedule (sec) & & $>1d$ & $0.10$ & $14.98$ & $1.64$\\
  $1$ & Alignment (sec) & & $>1d$ & $0.04$  & $0.75$ & $0.02$\\
  \cline{4-7}
  $8$ & Schedule (sec) & & $>1d$ & $0.20$ & $0.21$ & $0.30$ \\
  $8$ & Alignment (sec) & & $>1d$ & $0.12$ & $0.27$ & $0.02$ \\
  \cline{4-7}
  $1$ & Combined (sec) &  &$MEMOUT$ & $0.89$ & $32.09$ & $>1d$ \\
  \hline\hline
\end{tabular}
\caption{Runtimes of \clasp\ in CPU seconds to calculate an optimal schedule for one and eight cores.}
\label{tab:claspTime}
\end{table}
Table~\ref{tab:claspTime} shows the time 
spent on the computation \tplp{and the proof of the optimality} of timeout-minimal schedules and time-minimal alignments 
on the previously described benchmark sets 
for sequential schedules (first two rows) and parallel schedules for eight cores (next two rows).
\tplp{For the \satrand{}, \sathand{} and \cspset{} benchmark sets, 
the computation of the sequential and parallel schedule always took less
than one CPU second.
Some more time  
was spent for the \satindu{}, \qbfset{} and \maxsat{} benchmark set
but it is still feasible to find an optimal schedule.
We observe that the computation of parallel time slices is faster
than the computation of sequential schedules, 
except for the very simple \cspset{}.}
Given the additional processing units, 
the solvers can be scheduled more freely,
resulting in a less constrained problem that is easier to solve.
Furthermore, calculating a time-minimal alignment is easier in the parallel setting.
In our experiments, we obtained fewer selected solvers on the individual cores than in the sequential case.
This leads to smaller permutations of solvers and, in turn, reduces the total runtime.
\tplp{For the \aspset{}, we could not establish the optimal schedule even after one CPU day
and for the \SSSset{}, the calculation of the optimal schedule and optimal alignment was also impossible.
However \schedule{} was nevertheless able to find schedules and aligments,
and hence, was able to minimize the number of timeouts and runtime.
Finally, it is also possible that \schedule{} found an optimal schedule 
but was unable to prove its optimality.}
Therefore, we limited the maximal runtime of \clasp\ for these sets to $1200$ CPU seconds in all further experiments,
and used the resulting sub-optimal schedules and alignments obtained for this time.\footnote{
Note that
in our experiments,
the performance of \textit{unclasp}~\cite{ankamasc12a}, which optimizes based on unsatisfiable cores, did not exceed the perfomance of \clasp{} in computing solver schedules.
}

We also ran experiments on an encoding that optimizes the schedule and alignment simultaneously;
this approach accounts for all permutations of all potential timeout-minimal schedules.
The results are presented in the row labelled `Combined' in Table~\ref{tab:claspTime}.
The combination increases the solving time drastically.
Within one CPU day, 
\clasp{} was able to find an optimal solution and proved optimality only \tplp{for \sathand{}, \cspset{} and \qbfset{}}.
In all other cases, we aborted \clasp\ after one CPU day
and then used the best schedules found so far.
Nevertheless, we could find better alignments than in our two step approach 
(between $0.6\%$ and $9.8\%$ improvement), 
at the cost of substantially higher computation time and memory.
Because this encoding has a very large instantiation\tplp{, viz., more than $12$~GB memory consumption,} 
we were unable to run \schedule\ using it on the \SSSset{} and \aspset{}.
 

\subsection{Evaluation of Timeout-Minimal Schedules}
Having established that optimal schedules can be computed within a reasonable time in most cases, 
we evaluated the sequential timeout-minimal schedule of \schedule{} corresponding to the first step of our optimization process 
(cf. Equation~\eqref{eq:criterion:one}).
The number of timeouts for a fixed time budget assesses the robustness of a solver and 
is in many applications and competitions the primary evaluation criterion.

\begin{table}
\begin{tabular}{lrrrr}
 \hline\hline
   & \multicolumn{1}{c}{\satrand} & \multicolumn{1}{c}{\sathand} & \multicolumn{1}{c}{\satindu} & \multicolumn{1}{c}{\aspset}\\
 \cline{2-5}
 \sbs  			& $254/600$ 	& $155/300$ & $85/300$ 	& $446/2589$ \\
 \uni 			& $155/600$ 	& $123/300$ & $116/300$ & $536/2589$ \\
 \combi 		& $127/600$ 	& $126/300$ & $88/300$ 	& $308/2589$\\
 \selection  	& $\mathbf{115}/600$		& $101/300$ & $\mathbf{74}/300$  & $296/2589$\\
 \schedule  	& $131/600$ 	& $\mathbf{98}/300$  & $83/300$  & $\mathbf{290}/2589$\\
 \oracle  		& $108/600$ 	& $77/300$  & $45/300$  & $156/2432$\\
 \hline\hline
  & \multicolumn{1}{c}{\SSSset} & \multicolumn{1}{c}{\cspset} & \multicolumn{1}{c}{\qbfset} & \multicolumn{1}{c}{\maxsat}\\
 \cline{2-5}
 \sbs  			& $1881/5467$  	& $288/2024$ & $579/1368$ & $99/337$\\
 \uni 			& $1001/5467$  	& $283/2024$ & $357/1368$ & $21/337$\\
 \combi 		& $796/5467$  	& $283/2024$ & $357/1368$ & $10/337$\\
 \schedule  	& $\mathbf{603}/5467$ 	& $\mathbf{275}/2024$ & $\mathbf{344}/1368$ & $\mathbf{7}/337$\\
 \oracle  		& $0/5467$ 		& $253/2024$ & $314/1368$ & $0/337$\\
  \hline\hline  
\end{tabular}
\caption{Comparison of different approaches w.r.t. \#timeouts $/$ \#instances. 
The performance of the best performing system is in boldface.}
\label{tab:seqTO}
\end{table}

To obtain an unbiased evaluation of performance,
we used $10$-fold cross validation, a standard technique from machine learning:
First, the runtime data for a given instance set are randomly divided into $10$ equal parts.
Then, in each of the ten iterations, ${9}/{10}$th of the data is used as a training set for the computation of the schedule 
and the remaining ${1}/{10}$th serves as a test set to evaluate the performance of the solver schedule at hand;
\tplp{the results shown are obtained by summing over the folds.}
We compared the schedules computed by \schedule\ against the performance obtained 
from the \sbs, \uni, \combi, \selection{} (\satzilla{} and \claspfolio{}; if possible) approaches and the (theoretical) \oracle. 
The latter provides a bound on the best performance obtainable from any portfolio-based solver.

Table~\ref{tab:seqTO} shows the fraction of instances in each set on which timeouts occurred
(smaller numbers indicate better performance).
In all cases, \schedule\ showed better performance than the \sbs\ solver.
For example, \schedule\ reduced the number of timeouts from $1881$ to $603$ instances (less $23\%$ of unsolved instances) on the \SSSset,
\tplp{despite the fact that \schedule{} was unable to find the optimal schedule within the given $1200$ CPU seconds on this set.}
Also, \schedule\ performed better than the \uni\ approach.
The comparison with \combi\ and \selection{} (\satzilla{} and \claspfolio{}) revealed
that \schedule\ performed better than \combi\ in \tplp{seven out of eight scenarios we considered},
and better than \satzilla\ and \claspfolio{} in \tplp{two out of four scenarios}.
We expected that \schedule\ would solve fewer instances than the \selection{} approach in all four scenarios, 
because \schedule, unlike \satzilla{} and \claspfolio{}, does not use any instance features or prediction of solver performance.
It is somewhat surprising that \satzilla\ and \claspfolio{} do not always benefit from their more sophisticated approaches,
and further investigation into why this happens would be an interesting direction for future work.

\subsection{Evaluation of Time-Minimal Alignment}
After choosing the time slices for each solver, 
it is necessary to compute an appropriate solver alignment in order to obtain the best runtimes for our schedules.
As before, we used $10$-fold cross validation to assess this stage of \schedule.
To the best of our knowledge,
there is no system with a computation of alignments to compare against.
Hence,
we use a random alignment as a baseline
for evaluating our approach.
Thereby,
the expected performance of a random alignment is the average runtime of all possible alignments.
\tplp{Since the number of all permutations for \aspset{} and \SSSset{} is too large ($\gg 1\ 000\ 000\ 000$),
we approximate the performance of a random alignment by $10\ 000$ sampled alignments.} 

Table~\ref{tab:step2} shows the ratio of the expected performance of a random alignment 
and alignments computed by \schedule.
Note that
this ratio can be smaller than one,
because
the alignments are calculated on a training set and 
evaluated on a disjoint test set.

Also,
we contrast the optimal alignment with two easily computable heuristic alignments 
to avoid the search for an optimal alignment.
The alignment heuristic \heuO\ sorts solvers beginning with the solver with the minimal number of timeouts (most robust solver),
while \minO\ begins with the solver with the smallest time slice. 

\begin{table}
\begin{tabular}{lcccc}
  \hline\hline
   & \satrand & \sathand & \satindu & \aspset$^*$\\
  \cline{2-5}
  \schedule 	& $\mathbf{1.16}$ & $\mathbf{1.15}$ & $\mathbf{1.03}$ & $\mathbf{1.13}$\\
  \heuO 		& $1.02$ & $0.84$ & $1.00$ & $1.05$\\
  \minO 		& $1.15$ & $1.14$ & $1.00$ & $1.12$\\
  \hline\hline
  & \SSSset$^*$ & \cspset & \qbfset & \maxsat\\
  \cline{2-5}
  \schedule 	& $\mathbf{1.21}$ & $\mathbf{1.12}$ & $\mathbf{1.27}$ & $\mathbf{2.13}$ \\
  \heuO 		& $0.96$ & $0.90$ & $1.14$ & $0.89$ \\
  \minO 		& $1.20$ & $1.11$ & $1.14$ & $1.63$ \\
  \hline\hline
\end{tabular}
\caption{Ratios of the expected performance of a random alignment and alignments computed by \schedule{}, \heuO{} and \minO{}; 
\heuO{} sorts the solvers beginning with the solver with the minimal number of timeouts;
\minO{} begins with the solver with the smallest time slice.
The expected performance of a random alignment was approximated by $10.000$ samples for all sets marked with $^*$.}
\label{tab:step2}
\end{table}
As expected, the best performance is obtained by using optimal alignments within \schedule\ (Table~\ref{tab:step2});
it led, for example, to an increase in performance \tplp{by a factor of $2.13$ on \maxsat}.
In all cases, the performance of \minO\ was strictly better than (or equal to) that of \heuO.
Therefore, using \minO{} seems desirable whenever the computation of an optimal alignment is infeasible.

The actual runtimes of \schedule\ and the other approaches
are quite similar to the results on the number of timeouts (Table~\ref{tab:seqTO}) (data not shown).
The penalized runtimes (PAR$10$) are presented in Figure~\ref{fig:Trend} 
\subref{fig:Cores:Rand:PAR10},\subref{fig:Cores:Crafted:PAR10} and \subref{fig:Cores:Indu:PAR10}
at $\#cores=1$.

\subsection{Parallel Schedules}

\begin{figure}[h!]
  \centering
  \subfloat[\satrand\ - Parallel Schedule]{\label{fig:Cores:Rand:PAR10}\includegraphics[width=0.45\textwidth]{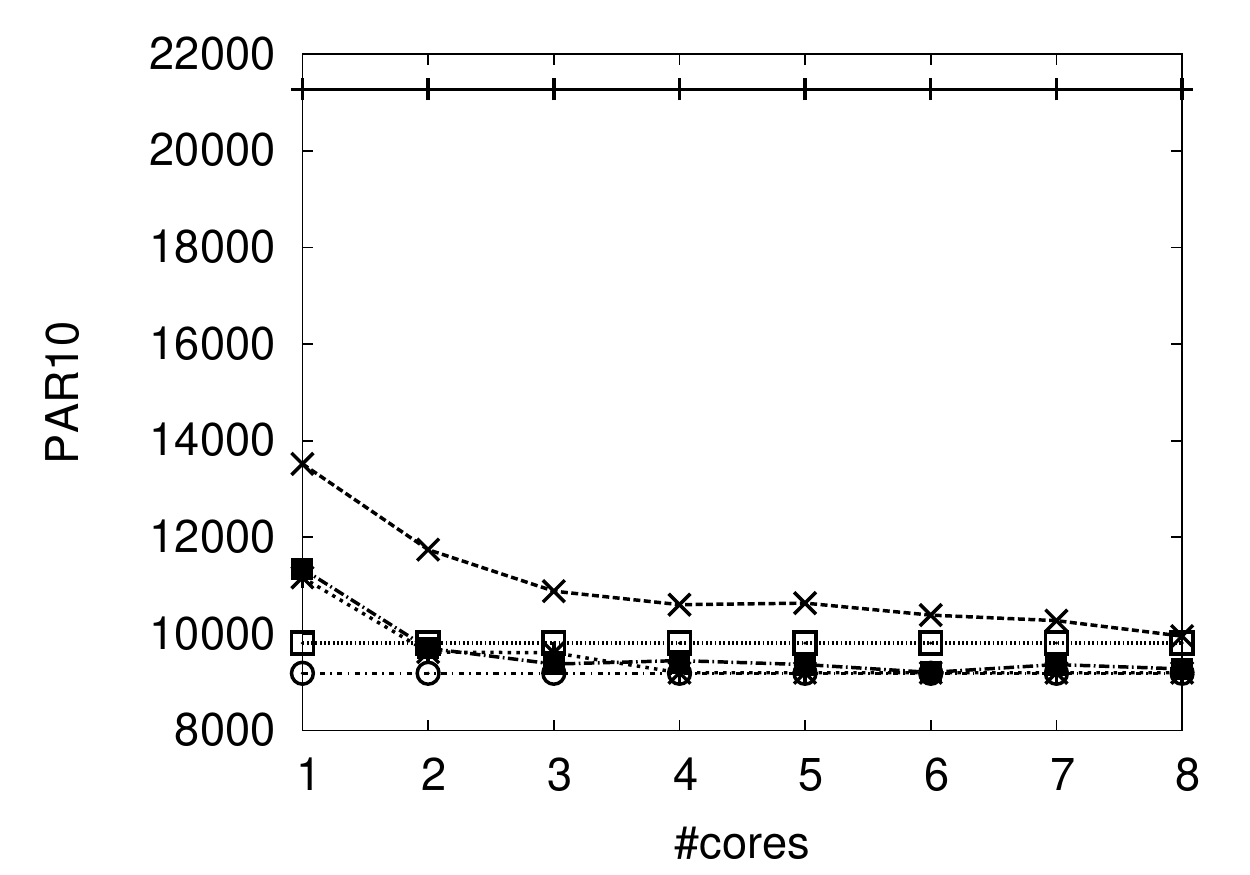}}     
  \subfloat[\satrand\ - Reduced Training Cutoff Time]{\label{fig:Cutoff:Rand:PAR10}\includegraphics[width=0.45\textwidth]{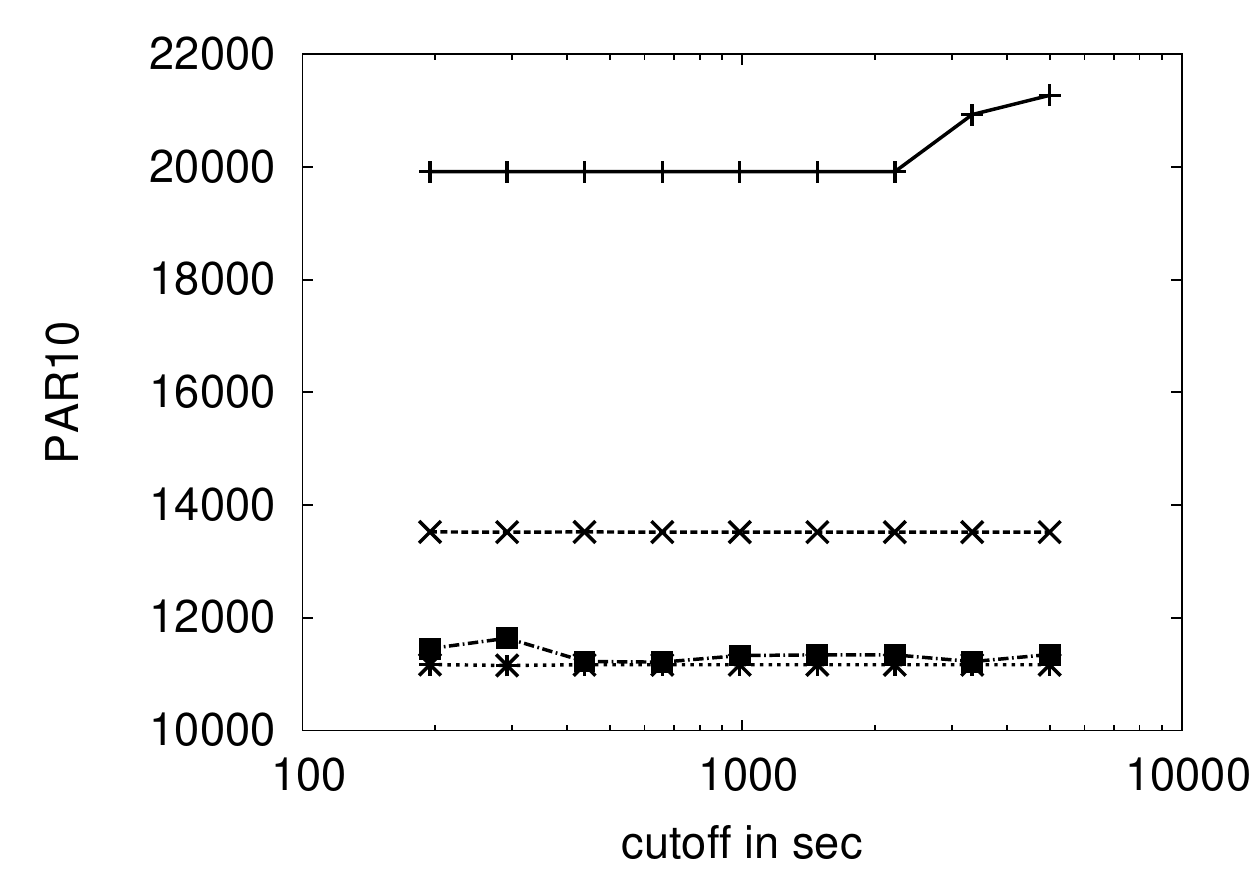}}
  
  \vspace*{-0.3cm}
  \subfloat[\sathand\ - Parallel Schedule]{\label{fig:Cores:Crafted:PAR10}\includegraphics[width=0.45\textwidth]{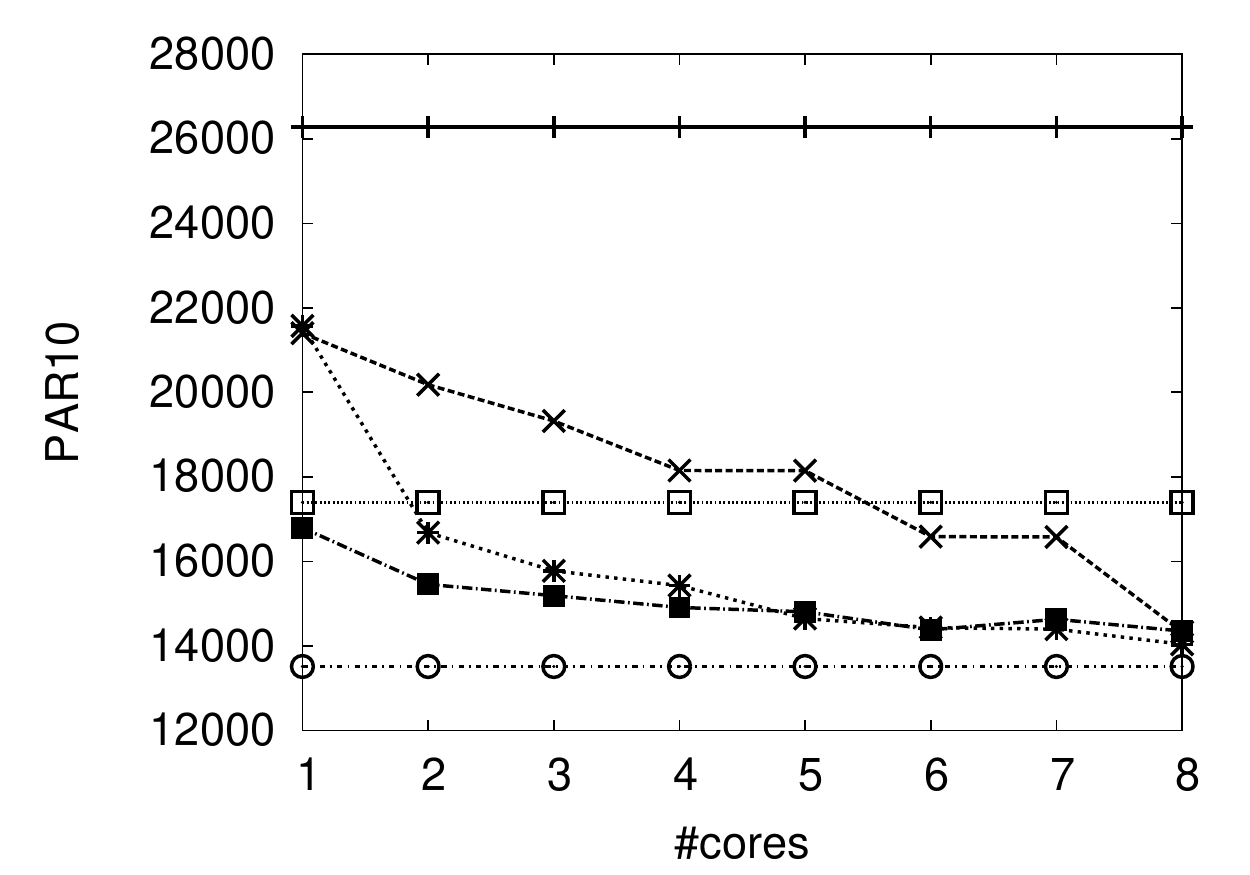}}
  \subfloat[\sathand\ - Reduced Training Cutoff Time]{\label{fig:Cutoff:Crafted:PAR10}\includegraphics[width=0.45\textwidth]{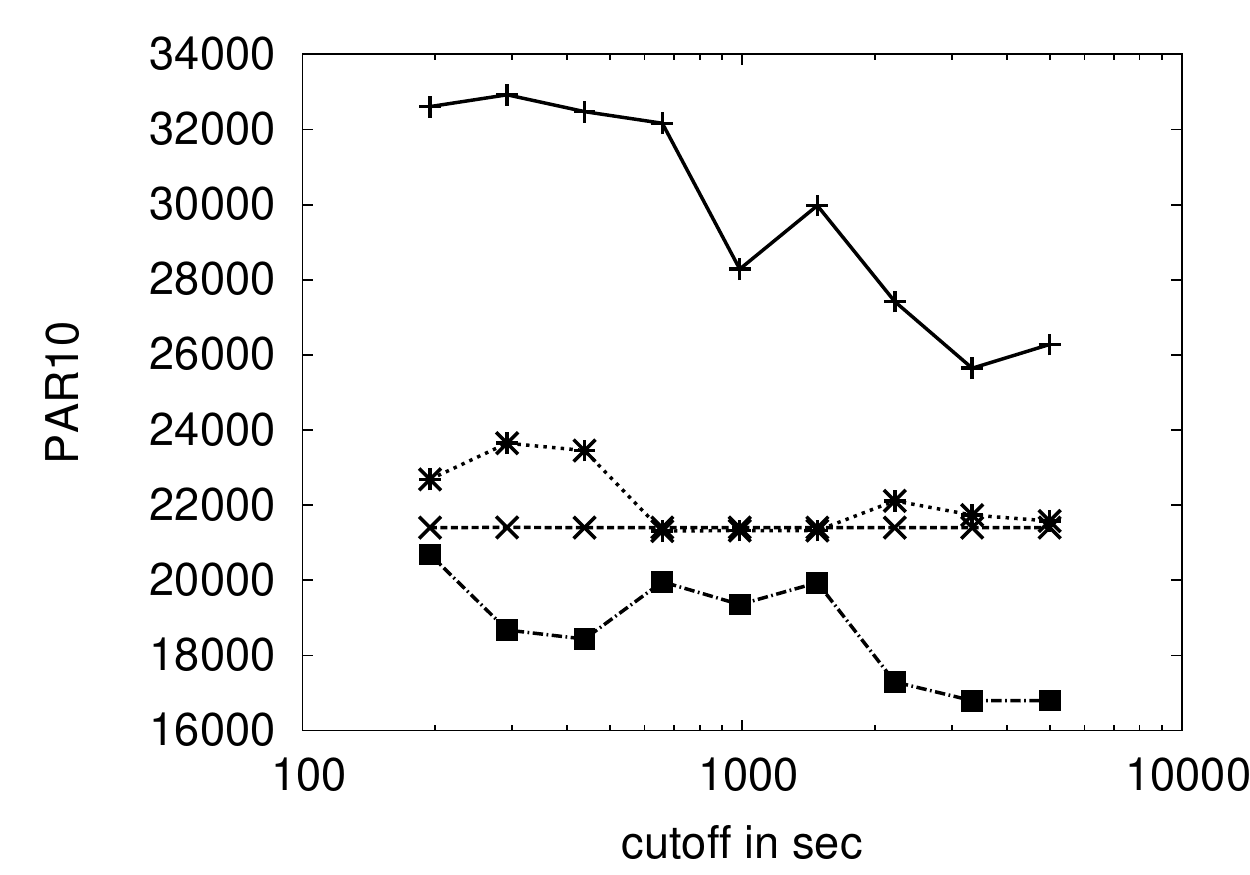}}
  
  \vspace*{-0.3cm}
  \subfloat[\satindu\ - Parallel Schedule]{\label{fig:Cores:Indu:PAR10}\includegraphics[width=0.45\textwidth]{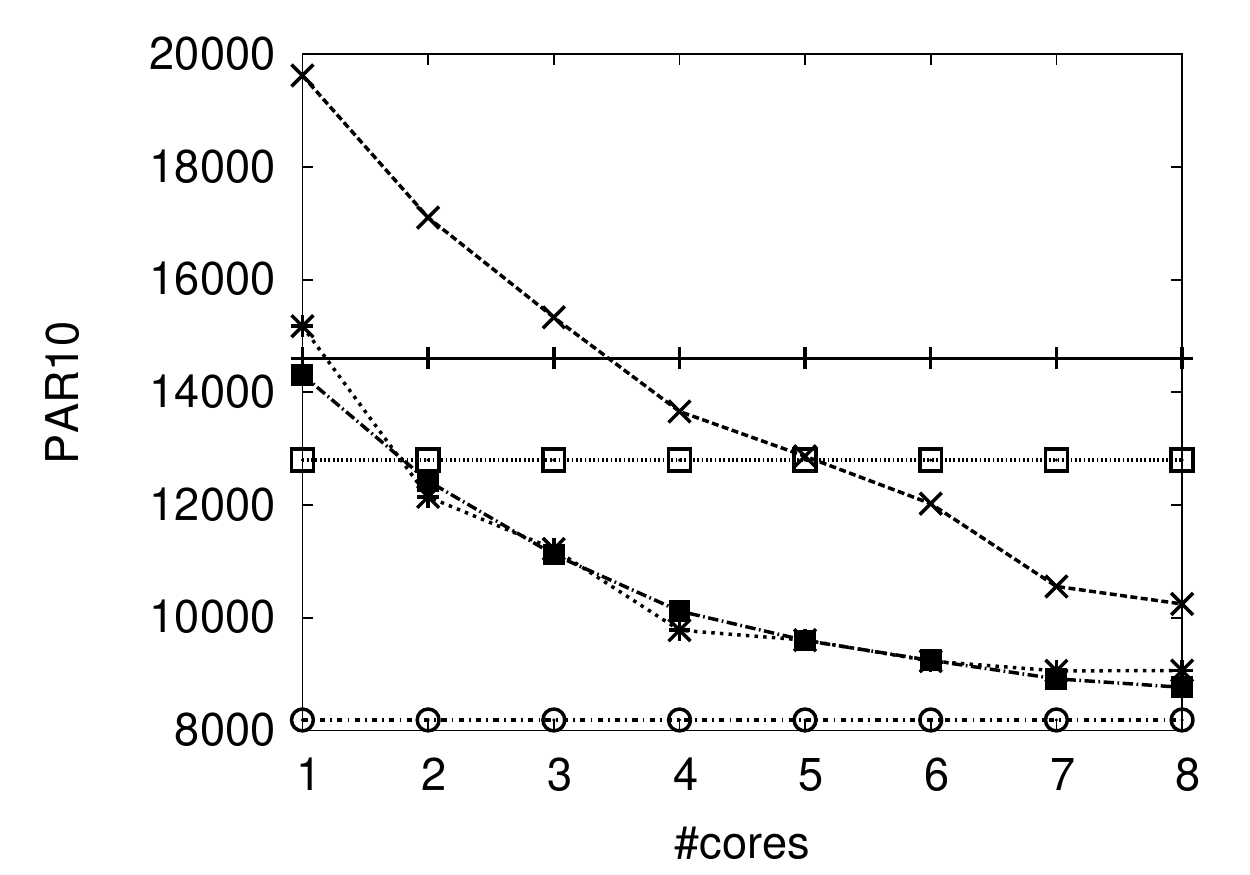}}
  \subfloat[\satindu\ - Reduced Training Cutoff Time]{\label{fig:Cutoff:Indu:PAR10}\includegraphics[width=0.45\textwidth]{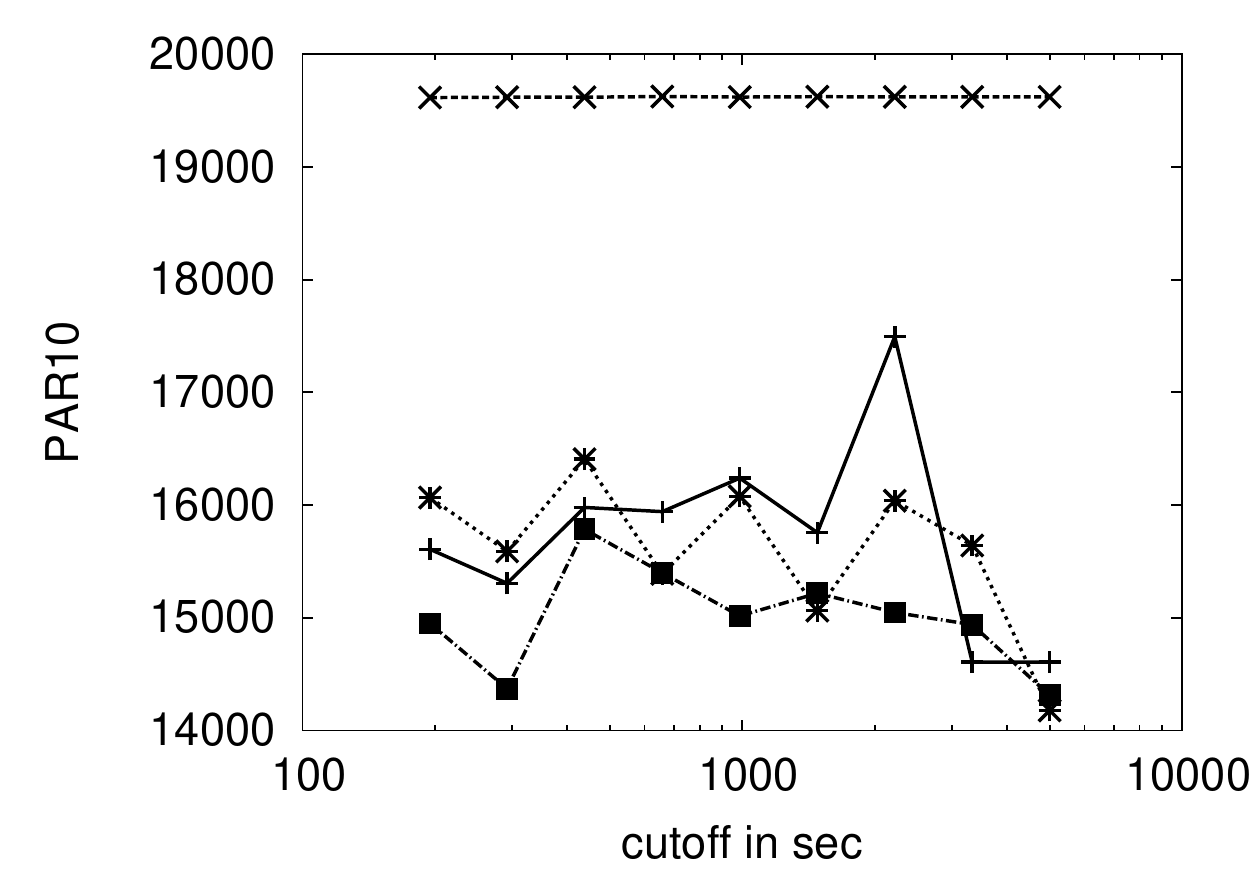}}
  
  \vspace*{-0.3cm}
  \subfloat[\aspset\ - Parallel Schedule]{\label{fig:Cores:ASP:PAR10}\includegraphics[width=0.45\textwidth]{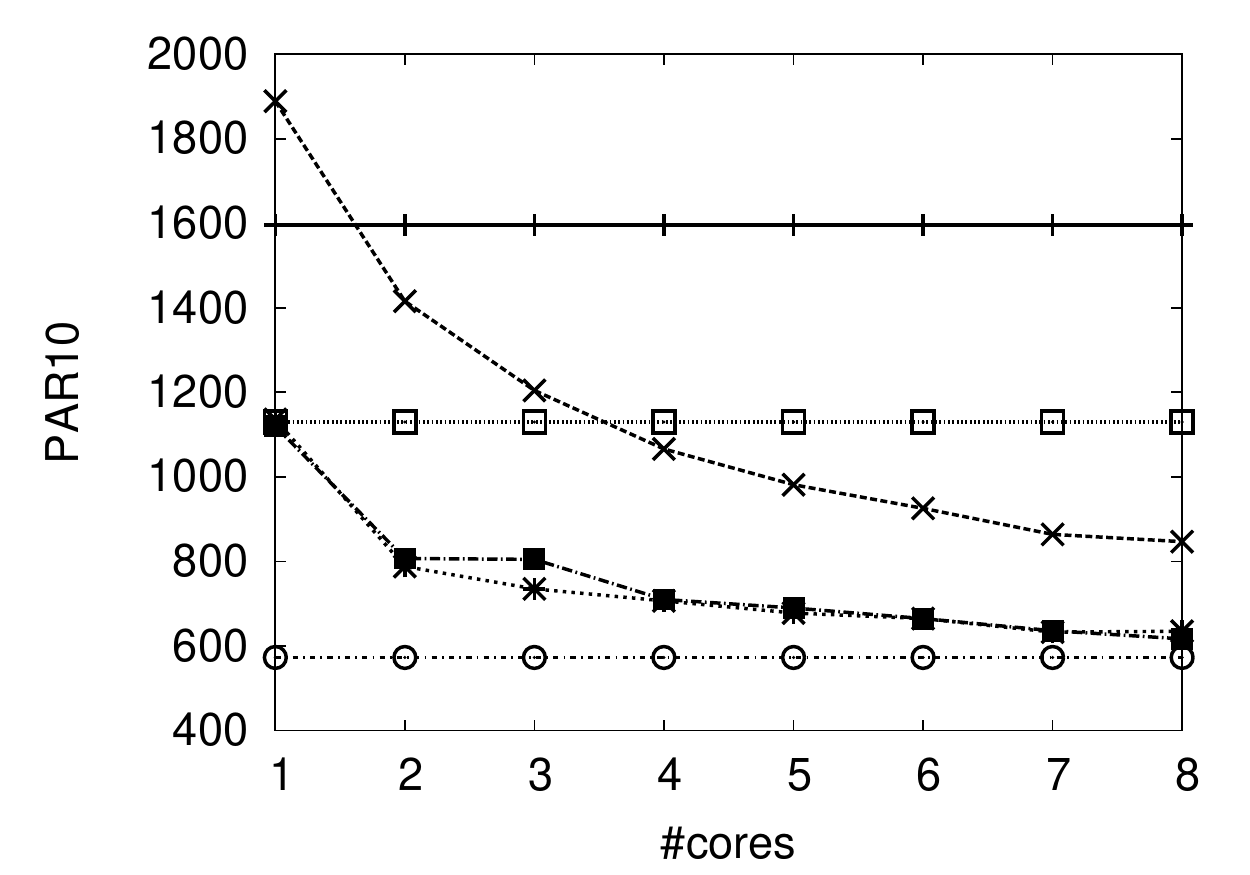}}
  \subfloat[\aspset\ - Reduced Training Cutoff Time]{\label{fig:Cutoff:ASP:PAR10}\includegraphics[width=0.45\textwidth]{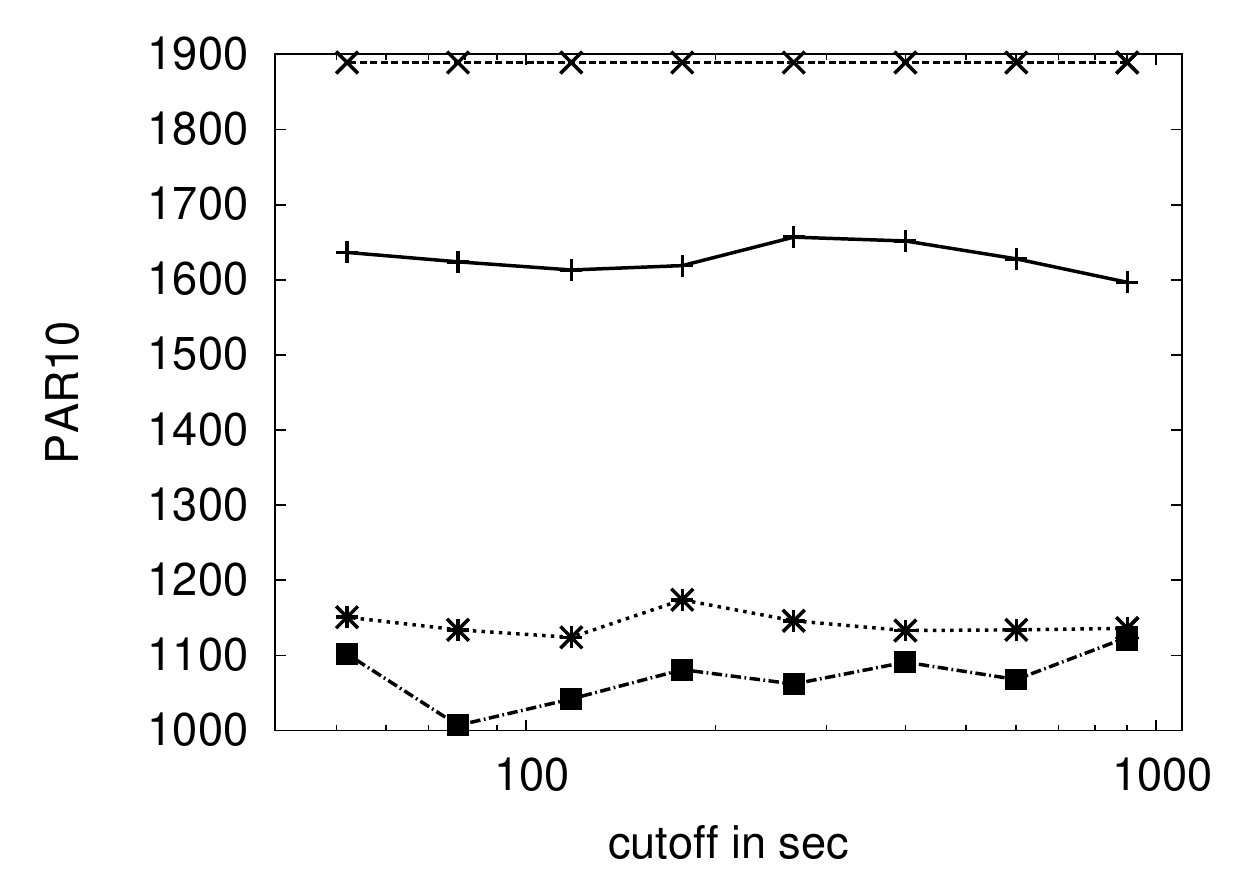}}
  
  \caption{Parallel Schedules (left) and reduced cutoff time (right), 
  			\sbs~($+$), \uni~($\times$), 
  			\combi\ approach~($\ast$), \schedule~($\blacksquare$), 
  			\selection~($\square$), \oracle~($\bigcirc$).}
  \label{fig:Trend}
\end{figure}

As we have seen in Section~\ref{sec:seq:asp},
our approach is easily extendable to parallel schedules.
We evaluated such schedules on \satrand{}, \sathand{}, \satindu{} \tplp{and \aspset{}}.
The results of this experiment are presented 
in Figure~\ref{fig:Trend} \subref{fig:Cores:Rand:PAR10}, \subref{fig:Cores:Crafted:PAR10}, 
\subref{fig:Cores:Indu:PAR10} \tplp{and \subref{fig:Cores:ASP:PAR10}}.
These evaluations were performed using $10$-fold cross validation and measuring wall-clock time.

In each graph, the number of cores is shown on the x-axis
and the PAR$10$ (penalized average runtime)\footnote{PAR$10$ penalizes each timeout with $10$ 
times the given cutoff time~\cite{huholest09a}.} on the y-axis;
we used PAR$10$, a commonly used metric from the literature, to capture average runtime as well as timeouts.
\third{(The sequential performance of \aspeed{} can be read off the values obtained for one core.)}
Since the \sbs\ solver~($+$) and \tplp{\selection~($\square$, \satzilla{} resp. \claspfolio)} cannot be run in parallel, 
their performance is constant.
\tplp{Furthermore, the \combi\ approach ($\ast$) is limited to run at most three component solver on the first core with uniform time slices
and one component solvers on each other core.
This more constrained schedule is also computed with the ASP encodings presented in Section~\ref{sec:seq:asp}
by adding three more constraints.}

As stated previously, the sequential version of \schedule~($\blacksquare$) performed worse than \satzilla~($\square$) in \satrand\ and \satindu.
However, \schedule\ turned out to perform at least as well as \satzilla\ when using two or more cores,
in terms of PAR$10$ scores as well in terms of average runtime (data not shown).
For example, \tplp{\scheduleMP{} -- that is parallel \schedule{} using four cores --} achieved a speedup 
of $1.20$ over the sequential \schedule\ 
on \satrand\ ($20$ fewer timeouts), $1.10$ on \sathand\ ($9$ fewer timeouts), $1.44$ on \satindu\ ($26$ fewer timeouts)
and \tplp{$1.57$ on \aspset{} ($111$ fewer timeouts)};
furthermore, \scheduleMP\ solved $4$, $13$, $17$\tplp{, $117$} instances more on these sets than (sequential) \satzilla{} \tplp{and \claspfolio{}, respectively}.
Considering the high performance of \satzilla~\cite{xuhuhole11b} \tplp{and \claspfolio~\cite{gekakascsczi11a}},
this represents a substantial performance improvement.

\begin{table}
\begin{tabular}{lrrrrrrrr}
 \hline\hline
  & \multicolumn{2}{c}{\SSSset} & \multicolumn{2}{c}{\cspset} & \multicolumn{2}{c}{\qbfset} & \multicolumn{2}{c}{\maxsat}\\
  & $\#$TO & PAR$10$ & $\#$TO  & PAR$10$ & $\#$TO  & PAR$10$ & $\#$TO  & PAR$10$\\
\hline
 \uni{}\textit{-SP}			& $1001$ & $9847$					& $283$ & $7077$ 					& $357$ & $10176$ 					& $21$ & $1470$\\
 \combi{}\textit{-SP}		& $796$  & $7662$					& $283$ & $7077$ 					& $357$ & $9657$ 					& $10$ & $731$\\
 \scheduleSP  				& $\textbf{603}$  & $\textbf{6001}$	& $\textbf{275}$ & $\textbf{6902}$ 	& $\textbf{344}$ & $\textbf{9272}$ 	& $\textbf{7}$ & $\textbf{516}$\\
 \hline
 \uni{}\textit{-4P}			& $583$  & $5720$					& $\textbf{253}$ & $\textbf{6344}$ 	& $\textbf{316}$ & $8408$ 			& $4$ & $511$\\
 \combi{}\textit{-4P} 		& $428$  & $4095$					& $\textbf{253}$ & $\textbf{6344}$ 	& $\textbf{316}$ & $8404$ 			& $4$ & $353$\\
 \scheduleMP  				& $\textbf{204}$  & $\textbf{2137}$ & $\textbf{253}$ & $\textbf{6344}$ 	& $\textbf{316}$ & $\textbf{8403}$ 	& $\textbf{3}$ & $\textbf{332}$\\
 \hline
 \oracle  					& $0$    & $198$ 					& $253$ & $6344$ & $314$ & $8337$ & $0$ & $39$\\
 \hline\hline
\end{tabular}
\caption{\tplp{Comparison of sequential and parallel schedules with $4$ cores w.r.t. \third{the number of timeouts and PAR$10$ score}.}}
\label{tab:parTO}
\end{table} 

Table~\ref{tab:parTO} presents the performance of parallel \aspeed{} with four cores (\scheduleMP),
\third{the parallel \uni{} and parallel \combi{} schedule, respectively,}
on \SSSset{}, \cspset{}, \qbfset{} and \maxsat{}.
We decided to use only four cores
because (i) \cspset{} and \qbfset{} have two resp. five solvers,
and therefore it is trivial to perform as well as the \oracle{} with $4$ or more cores,
and (ii) we saw in Figure~\ref{fig:Trend} that
the curves flatten beginning with four cores,
which is an effect of the complementarity of the solvers in the portfolio.
The performance of \scheduleSP{}, i.e., sequential \aspeed{}, 
is already nearly as good as the \oracle{} on \maxsat{}
and \scheduleMP{} was only able to improve the performance slightly.
However,
\scheduleMP{} was able to decrease the number of timeouts from $603$ to $204$ on the \SSSset{}.

\subsection{Generalization Ability of \schedule{}}
The schedule computation of \schedule\ uses runtime data measurements, 
which require extensive computational resources.
Therefore, we investigated
the possibility to decrease the cutoff time on the training data
to reduce the overall computational burden of training.
The schedules thus obtained were evaluated on test data with an unreduced cutoff time.
We note that
only instances are considered for the computation of schedules
that are solved by at least one solver in the portfolio.
Therefore, using this approach with a lower training cutoff time, 
the computation of a schedule is based on easier and fewer instances than 
those in the test set \edhh{used to ultimately evaluate it}.
Figures \ref{fig:Trend} \subref{fig:Cutoff:Rand:PAR10}, \subref{fig:Cutoff:Crafted:PAR10} and \subref{fig:Cutoff:Indu:PAR10}
show the results of evaluating the resulting schedules in the same way 
as in the experiments for parallel schedules with $10$-fold cross validation \tplp{but using only one processing unit}.
The cutoff time on the training set (shown on a logarithmic x-axis) was reduced according to a ${2}/{3}$-geometric sequence,
from the maximal cutoff time of $5000$ down to $195$ CPU seconds \tplp{for \satrand{}, \sathand{} and \satindu{}
and $900$ down to $52$ CPU seconds for the \aspset{}.
}
A flat line corresponds to the expected optimal case
that the performance of a schedule does not suffer from a reduced cutoff time;
the \uni\ approach ($\times$) does not rely on training data and 
therefore has such a constant performance curve.

Surprisingly, the reduced cutoff time had nearly no effect on the performance of \schedule~($\blacksquare$) 
on \satrand\ (Figure~\ref{fig:Cutoff:Rand:PAR10}) \tplp{and \aspset{} (Figure~\ref{fig:Cutoff:ASP:PAR10})}.
On the other hand, the selection of the \sbs\ solver ($+$) got worse with an increased cutoff time on the training data \tplp{of \satrand{}}.
On the \sathand\ set (Figure~\ref{fig:Cutoff:Crafted:PAR10}), the performance of \schedule\ was found to benefit from an increased cutoff time,
but the improvement was small for a cutoff time longer than $2222$ CPU seconds (${4}/{9}$ of the maximal cutoff time).
In contrast, the improvement of the \combi\ approach ($\ast$) was small on \sathand\ and \satrand;
and the performance of \schedule, \combi\ approach and \sbs\ fluctuated on the \satindu\ set (Figure~\ref{fig:Cutoff:Indu:PAR10}).
All three approaches benefited from the maximal cutoff time ($5000$ CPU seconds);
however, the benefit was small in comparison to \schedule\ with the fully reduced cutoff time ($195$ CPU seconds).
We conjecture that in the case of \sathand, the easy instances are not representative for the harder instances in the test set, unlike in the case of \satrand, \edhh{where} all instances were randomly generated and
of similar structure. 
Consequently, on sets like \satrand, easier instances can be used for the computation of a schedule,
even if the resulting schedule is ultimately applied to (and evaluated on) harder instances.

\begin{table}
\begin{tabular}{lcccc}
   \hline\hline
   & \satrand & \sathand & \satindu & \textit{Complete} \\
  \cline{2-5}
  \sbs & $23662$ & $29906$ & $16942$ & $32457$\\
  \schedule & $19061$ & $24623$ & $16942$ & $21196$\\
  \hline\hline
\end{tabular}
\caption{PAR10 of \sbs{} and \schedule{}, trained on $2009$ SAT Competition and evaluated on $2011$ SAT Competition.}
\label{tab:competition}
\end{table}

In an additional experiment, we assessed the performance of \schedule{}
in the context of preparing for a competition.
\schedule\ was trained on instances of the $2009$ SAT Competition with the SAT solvers
\cryptominisat, \clasp\ and \tnm,
which are the same solvers used by \ppfolio, 
and evaluated on the instances of the $2011$ SAT Competition; see Table~\ref{tab:competition}.
On the entire instance set, \schedule\ had a PAR$10$  of $21\,196$, in contrast to the \sbs\ solver with $32\,457$ (a factor $1.53$ higher).
Also, \schedule\ outperformed the \sbs\ solver on \satrand\ and \sathand, 
and it performed just as well as the \sbs\ solver on \satindu.
This latter observation is due the fact that the performance of \cryptominisat{} dominated on the \satindu{} set,
and hence, \schedule{} was unable to obtain improved performance on \satindu{}.

\subsection{Comparison with \SSS}

In our final experiment, 
we compared \schedule{} with the SAT solver \SSS{},
which uses an approach similar to \schedule{},
but combines a static solver schedule with algorithm selection based on instance features (see Section \ref{sec:related:work}).
Since only \tplp{the sequential version of the solver \SSS{}} is freely available 
but not the schedule building method,
we could not train the models of \SSS{} on new benchmark sets.
Therefore, we trained \schedule{} on the same training runtime measurements 
used by the authors of \SSS{} for training on the $2011$ SAT Competition, namely the \SSSset{}.
We note that training of \SSS{}, unlike \schedule, additionally requires a set of instance features.
Using these versions of \schedule{} and \SSS{} trained on the same set of instances,
we measured the runtime of both solvers
(utilizing a single processor \textit{SP} or multi-processor environment with four parallel threads \textit{MP4})
on the instances of the $2011$ SAT Competition 
with the same cutoff of $5000$ CPU seconds as used in the competition.

\begin{table}
\begin{tabular}{lcccc}
  \hline\hline
   & \satrand & \sathand & \satindu & \textit{Complete} \\
  \cline{2-5}
  \SSS{} & $16415$ & $23029$ & $19817$ & $18919$\\
  \scheduleSP{} & $22095$ & $22180$ & $24579$ & $22737$\\
  \scheduleMP{} & $16380$ & $20142$ & $17164$ & $17517$\\
  \hline\hline
\end{tabular}
\caption{\edhh{PAR10 of \SSS{} and \schedule{}, trained on the training data of \SSS\ and evaluated on $2011$ SAT Competition.}}
\label{tab:competition2}
\end{table}

Table \ref{tab:competition2} shows the results based on the PAR$10$ of the runtime measurements.
The results are similar to the comparison between \satzilla{} and \schedule{}.
The single processor version of \schedule{}, \scheduleSP{}, outperformed \SSS{} on \sathand{} in the sequential case.
This could indicate 
that the instance feature set, used by \satzilla{} and \SSS{}, does not sufficiently reflect
the runtime behaviour of the individual solvers on these types of instances.
Furthermore, \schedule{} with four cores, \scheduleMP{}, performed better than \SSS{}
on all three instance sets.

%


\section{Related Work}
\label{sec:related:work}

Our work forms part of a long line of research that can be traced back to John Rice's seminal 
work on algorithm selection \citeyear{rice76a} on one side, and to work by Huberman, Lukos, and Hogg 
\citeyear{huluho97a} on parallel algorithm portfolios on the other side.

Most recent work on algorithm selection is focused on mapping problem instances 
to a given set of algorithms, where the algorithm to be run on a given problem instance $i$ is typically 
determined based on a set of (cheaply computed) features of $i$.
This is the setting considered prominently in \cite{rice76a},
as well as by the work on SATzilla, 
which makes use of regression-based models of running time \cite{xuhole07a,xuhuhole08a};
work on the use of decision trees and case-base reasoning for selecting bid evaluation algorithms in combinatorial auctions \cite{guemil04a,geguhnmi04a}; 
and work on various machine learning techniques 
for selecting algorithms for finding maximum probable explanations in Bayes nets in real time \cite{guohsu04}.
All these approaches are similar to ours in that they exploit complementary strengths of a set of solvers
for a given problem; however, unlike these per-instance algorithm selection methods, 
\schedule{} selects and schedules solvers to optimize performance on a set of
problem instances, and therefore does not require instance features.

It may be noted that the use of pre-solvers in \satzilla{}, i.e., solvers that are run
feature-extraction and feature-based solver selection, bears some resemblance to the sequential
solver schedules computed by \schedule{}; however, \satzilla{} considers only up to 2 pre-solvers,
which are determined based on expert knowledge (in earlier versions of SATzilla) or by exhaustive search, 
along with the time they are run for. 

\cphydra{} is a portfolio-based procedure for solving constraint programming problems 
that is based on case-based reasoning for solver selection and a simple complete search procedure
for sequential solver scheduling \cite{mahehonusu08a}.
Like the previously mentioned approaches, and unlike \schedule{}, it requires instance features
for solver selection, and, according to its authors, is limited to a low number of 
solvers (in their work, five). Like the simplest variant of \schedule{}, the solver scheduling in \cphydra{} 
aims to maximize the number of given problem instances solved within a given time budget.

Early work on parallel algorithm portfolios highlights the potential for performance improvements,
but does not provide automated procedures for selecting the solvers to be run in parallel from a larger
base set \cite{huluho97a,gomsel01a}. 
\ppfolio{}, which demonstrated impressive performance at the 2011 SAT Competition, 
is a simple procedure that runs between 3 and 5 SAT solvers concurrently (and, depending on the number of processors or cores available, potentially in parallel) on a given SAT instance. The component solvers have been chosen manually based on performance on past competition instances, and they are all run for the same amount of time.
Unlike \ppfolio, our approach automatically selects solvers to minimize the number
of timeouts or total running time on given training instances using a powerful ASP solver
and can, at least in principle, work with much larger numbers of solvers.
Furthermore, unlike \ppfolio{}, \schedule{} can allot variable amounts of time to each solver 
to be run as part of a sequential schedule.

Concurrently with our work presented here, Yun and Epstein \citeyear{yuneps12a} developed an approach 
that builds sequential and parallel solver schedules using case-based
reasoning in combination with a greedy construction procedure.
Their RSR-WG procedure combines fundamental aspects of \cphydra{} \cite{mahehonusu08a}
and GASS \cite{stgosm07a}; unlike \schedule{}, it relies on instance features.
RSR-WG uses a relatively simple greedy heuristic 
to optimize the number of problem instances solved within a given time budget
by the parallel solver schedule to be constructed;
our use of an ASP encoding, on the other hand, offers considerably more flexibility in formulating
the optimization problem to be solved, and our use of powerful, general-purpose ASP solvers
can at least in principle find better schedules. Our approach also goes beyond
RSR-WG in that it permits the optimization of parallel schedules for runtime.

Gagliolo and Schmidhuber consider a different setting, in which a set of algorithms is
run in parallel, with dynamically adjusted timeshares \citeyear{gagsch06a}.
They use a multi-armed bandit solver to allocate timeshares to solvers
and present results using two algorithms for SAT and winner determination in combinatorial
auctions, respectively. Their technique is interesting, but considerably more complex than
\schedule{}; while the results for the limited scenarios they studied are promising, so far, there
is no indication that it would achieve state-of-the-art performance
in standardized settings like the SAT competitions.

\tplp{For AI planning,
Helmert et al.\ implemented the portfolio solver \stonesoup{}~\cite{heroka11a,sebrgahe12a}
which statically schedules planners.
In contrast to \schedule{},
\stonesoup{}s computes time slices using a greedy hill climbing algorithm
that optimizes a special planning performance metric,
and the solvers are aligned heuristically.
The results reported by \citeN{sebrgahe12a} showed that
an uniform schedule achieved performance superior to that of \stonesoup.
Considering our results about uniform schedules and schedules computed by \aspeed{},
we have reason to believe
that the schedules optimized by \aspeed{} could also achieve performance improvements on AI planning problems.
}

Perhaps most closely related to our approach is the recent work of Kadioglu et al.\ on algorithm selection
and scheduling \cite{kamasasase11a}, namely \SSS. They study pure algorithm selection and various 
scheduling procedures based on mixed integer programming techniques.
Unlike \schedule{}, their more sophisticated procedures rely on instance features for nearest-neighbour-based
solver selection, based on the (unproven) 
assumption that any given solver shows similar performance on instances with similar features 
\cite{kamaseti10a}. 
\tplp{(We note that solver performance is known to vary substantially over sets of artificially created, 
‘uniform random’ SAT and CSP instances that are identical in terms of cheaply computable syntactic features, 
suggesting that this assumption may in fact not hold.)}
\tplp{The most recent version of \SSS{}~\cite{masasase12a} also supports the computation of parallel schedules
but is unfortunately not available publicly or for research purposes.}
We focussed deliberately on a simpler setting than their best-performing semi-static scheduling approach in that we
do not use per-instance algorithm selection, yet still obtain excellent performance.
\tplp{Furthermore, \SSS{} only optimizes the number of timeouts
whereas \schedule{} also optimizes the solver alignment to improve the runtime.}



\section{Conclusion}
\label{sec:conclusion}

In this work, we demonstrated how ASP formulations and a powerful ASP solver (\clasp) can be used
to compute sequential and parallel solver schedules. 
\third{In principle, a similar approach could be pursued using CP or ILP as done within \SSS{}~\cite{kamasasase11a}.
However, as we have shown in this work, 
ASP appears to be a good choice,
since it allows for a compact and flexible encoding of the specification, 
for instance, by supporting true multi-objective optimization, 
and can be applied to effectively solve the problem for many domains.}

Compared to earlier model-free and 
model-based approaches (\ppfolio{} and \satzilla{}, respectively), 
our new procedure, \schedule{}, performs very well on \tplp{ASP, CSP, MaxSAT, QBF and SAT} -- five widely studied 
problems for which substantial and sustained effort is being expended in the design
and implementation of high-performance solvers.
In the case of SAT, there is no single dominant solver, and portfolio-based approaches
leverage the complementary strength of different state-of-the-art algorithms.
For ASP, a situation exists with respect to different configurations of a single 
solver, \clasp{}. 
This latter case is interesting, because we essentially use \clasp{} to optimize itself.
While, in principle, the kind of schedules 
we construct over various configurations of \clasp{} could even be used
\emph{within} \schedule{} instead of plain \clasp{}, we have not yet investigated the
efficacy of this approach.

Our open-source reference implementation of \schedule{} is available online.
We expect \schedule{} to work particularly well in situations where various different kinds 
of problem instances have to be solved (e.g., competitions) or where single good 
(or even dominant) solvers or solver configurations are unknown (e.g., new applications).
Our approach leverages the power of multi-core and multi-processor 
computing environments and, because of its use of easily modifiable and extensible ASP encodings,
can in principle be readily modified to accommodate different constraints on 
and optimization criteria for the schedules to be constructed.
Unlike most other portfolio-based approaches, \schedule{} does not require instance features
and can therefore be applied more easily to new problems.

Because, like various other approaches, \schedule{} is based on minimization of timeouts, 
it is currently only applicable in situations where some instances cannot be solved
within the time budget under consideration (this setting prominently arises in many 
solver competitions). In future work, we intend to investigate strategies that automatically
reduce the time budget if too few timeouts are observed on training data;
we are also interested in the development of better techniques for directly minimizing runtime.

In situations where there is a solver or configuration that dominates all others 
across the instance set under consideration, portfolio-based approaches are generally not effective
(with the exception of performing multiple independent runs of a randomized solver).
The degree to which performance advantages can be obtained through the use of 
portfolio-based approaches, and in particular \schedule{}, depends on the degree to which
there is complementarity between different solvers or configurations, and it would be 
interesting to investigate this dependence quantitatively, possibly based on 
recently proposed formal definitions of instance set homogeneity \cite{hoosch11a}.
\third{Alternatively, 
if a dominant solver configuration is expected to exist
but is unknown,
such a configuration could be found using an algorithm configurator,
for instance \textit{ParamILS}~\cite{huhost07a,huholest09a}, \textit{GGA}~\cite{anseti09a}, \textit{F-Race}~\cite{lodustbi11a} or \textit{SMAC}~\cite{huhole11b}.
Furthermore, automatic methods, like \hydra{}~\cite{xuhole10a} and \isac{}~\cite{kamaseti10a}, 
construct automatically complementary portfolios of solver configurations with the help of algorithm configurators
which could be also combined with \aspeed{} to further increase its performance.
}

\paragraph*{Acknowledgments}
This work was partially funded by the German Science Foundation (DFG) under
grant SCHA 550/8-3.

 \bibliographystyle{acmtrans} 

\begin{thebibliography}{}

\end{thebibliography}


\begin{thebibliography}{}

\bibitem[\protect\citeauthoryear{Butcher}{Butcher}{1981}]{Butcher}
{\sc Butcher, J.} 1981.
\newblock {\em Copy-editing: The Cambridge Handbook}.
\newblock Cambridge University Press.

\bibitem[\protect\citeauthoryear{{{C}adence {R}esearch {S}ystems}}{{{C}adence
  {R}esearch {S}ystems}}{1994}]{crs:chez}
{\sc {{C}adence {R}esearch {S}ystems}}. 1994.
\newblock {\it {C}hez} {S}cheme {R}eference {M}anual.

\bibitem[\protect\citeauthoryear{Cameron and Ito}{Cameron and
  Ito}{1984}]{ci:gramps}
{\sc Cameron, R.~D.} {\sc and} {\sc Ito, M.~R.} 1984.
\newblock Grammar-based definition of metaprogramming systems.
\newblock {\em {ACM} Transactions on Programming Languages and Systems\/}~{\em
  6,\/}~1 (Jan.), 20--54.

\bibitem[\protect\citeauthoryear{Grossman}{Grossman}{1982}]{Chicago}
{\sc Grossman, J.}, Ed. 1982.
\newblock {\em The Chicago Manual of Style}.
\newblock University of Chicago Press.

\bibitem[\protect\citeauthoryear{Lamport}{Lamport}{1986}]{LaTeX}
{\sc Lamport, L.} 1986.
\newblock {\em \LaTeX: A Document Preparation System\/}, 2 ed.
\newblock Addison-Wesley, New York.

\end{thebibliography}


\begin{thebibliography}{}

\bibitem[\protect\citeauthoryear{Apt and Bol}{Apt and Bol}{1994}]{apt:94}
{\sc Apt, K.} {\sc and} {\sc Bol, R.} 1994.
\newblock Logic programming and negation: A survey.
\newblock {\em Journal of Logic Programming\/}~{\em 19,20}, 9--71.

\bibitem[\protect\citeauthoryear{Brewka}{Brewka}{1996}]{brewka:plp}
{\sc Brewka, G.} 1996.
\newblock Well-founded semantics for extended logic programs with dynamic
  preferences.
\newblock {\em Journal of Artificial Intelligence Research\/}~{\em 4}, 19--36.

\bibitem[\protect\citeauthoryear{Brewka and Eiter}{Brewka and
  Eiter}{1999}]{be:aij}
{\sc Brewka, G.} {\sc and} {\sc Eiter, T.} 1999.
\newblock Preferred answer sets for extended logic programs.
\newblock {\em Artificial Intelligence\/}~{\em 109}, 297--356.

\bibitem[\protect\citeauthoryear{Choelwinski}{Choelwinski}{1994}]{ch:stra}
{\sc Choelwinski, P.} 1994.
\newblock Stratified default logic.
\newblock In {\em Proceedings of Computer Science Logic}. Springer, LNCS 933,
  456--470.

\bibitem[\protect\citeauthoryear{Das}{Das}{1992}]{lp:92}
{\sc Das, S.} 1992.
\newblock {\em Deductive Database and Logic Programming}.
\newblock Addison-Wesley Publishers Ltd.

\bibitem[\protect\citeauthoryear{Delgrande, Schaub, and Tompits}{Delgrande
  et~al\mbox{.}}{2000}]{d:plp}
{\sc Delgrande, J.}, {\sc Schaub, T.}, {\sc and} {\sc Tompits, H.} 2000.
\newblock Logic programs with complied preferences.
\newblock In {\em Proceedings of the 14th European Conference on Artificial
  Intelligence (ECAI-2000)}. 392--398.

\bibitem[\protect\citeauthoryear{Gelfond and Lifschitz}{Gelfond and
  Lifschitz}{1988}]{gl:stable}
{\sc Gelfond, M.} {\sc and} {\sc Lifschitz, V.} 1988.
\newblock The stable model semantics for logic programming.
\newblock In {\em Proceedings of the Fifth Joint International Conference and
  Symposium}. MIT Press, 1070--1080.

\bibitem[\protect\citeauthoryear{Gelfond and Lifschitz}{Gelfond and
  Lifschitz}{1991}]{gl:logic}
{\sc Gelfond, M.} {\sc and} {\sc Lifschitz, V.} 1991.
\newblock Classical negation in logic programs and disjunctive databases.
\newblock {\em New Generation Computing\/}~{\em 9}, 365--386.

\bibitem[\protect\citeauthoryear{Gelfond and Son}{Gelfond and
  Son}{1998}]{ms:98}
{\sc Gelfond, M.} {\sc and} {\sc Son, T.} 1998.
\newblock Reasoning with prioritized defaults.
\newblock In {\em LNAI 1471}. Springer, 164--224.

\bibitem[\protect\citeauthoryear{Grosof}{Grosof}{1997}]{g:97}
{\sc Grosof, B.} 1997.
\newblock Prioritized conflict handling for logic programs.
\newblock In {\em Proceedings of the 1997 International Logic Program Symposium
  (ILPS'97)}. MIT Press, 197--212.

\bibitem[\protect\citeauthoryear{Lifschitz and Turner}{Lifschitz and
  Turner}{1994}]{lt:lp94}
{\sc Lifschitz, V.} {\sc and} {\sc Turner, H.} 1994.
\newblock Splitting a logic program.
\newblock In {\em Proceedings of Eleventh International Conference on Logic
  Programming}. MIT Press, 23--37.

\bibitem[\protect\citeauthoryear{Schaub and Wang}{Schaub and
  Wang}{2001}]{sw:01}
{\sc Schaub, T.} {\sc and} {\sc Wang, K.} 2001.
\newblock A comparative study of logic programs with preference.
\newblock In {\em Proceedings of the 17th International Joint Conference on
  Artificial Intelligence (IJCAI-2001)}. Morgan Kaufmann Publishers Inc.,
  597--602.

\bibitem[\protect\citeauthoryear{Wang, Zhou, and Lin}{Wang
  et~al\mbox{.}}{2000}]{w:plp}
{\sc Wang, K.}, {\sc Zhou, L.}, {\sc and} {\sc Lin, F.} 2000.
\newblock Alternating fixpoint theory for logic programs with priority.
\newblock In {\em Proceedings of International Joint Conference on
  Computational Logic (CL-2000)}. 164--178.

\bibitem[\protect\citeauthoryear{Zhang}{Zhang}{1999}]{yan:iclp99}
{\sc Zhang, Y.} 1999.
\newblock Monotonicity in rule based update.
\newblock In {\em Proceedings of the 1999 International Conference on Logic
  Programming (ICLP'99)}. MIT Press, 471--485.

\bibitem[\protect\citeauthoryear{Zhang}{Zhang}{2001}]{yan:00lp}
{\sc Zhang, Y.} 2001.
\newblock The complexity of logic program update.
\newblock In {\em Proceedings of the 14th Australian Joint Conference on
  Artificial Intelligence (AI2001)}. Springer, LNAI 2256, 630--643.

\bibitem[\protect\citeauthoryear{Zhang and Foo}{Zhang and Foo}{1997}]{yan:plp}
{\sc Zhang, Y.} {\sc and} {\sc Foo, N.} 1997.
\newblock Answer sets for prioritized logic programs.
\newblock In {\em Proceedings of the 1997 International Logic Programming
  Symposium (ILPS'97)}. MIT Press, 69--83.

\end{thebibliography}


\begin{thebibliography}{}

\bibitem[\protect\citeauthoryear{Andres, Kaufmann, Matheis, and Schaub}{Andres
  et~al\mbox{.}}{2012}]{ankamasc12a}
{\sc Andres, B.}, {\sc Kaufmann, B.}, {\sc Matheis, O.}, {\sc and} {\sc Schaub,
  T.} 2012.
\newblock Unsatisfiability-based optimization in clasp.
\newblock In {\em Technical Communications of the Twenty-eighth International
  Conference on Logic Programming (ICLP'12)}, {A.~Dovier} {and} {V.~{Santos
  Costa}}, Eds. Vol.~17. Leibniz International Proceedings in Informatics
  (LIPIcs), 212--221.

\bibitem[\protect\citeauthoryear{Ans{\'o}tegui, Sellmann, and
  Tierney}{Ans{\'o}tegui et~al\mbox{.}}{2009}]{anseti09a}
{\sc Ans{\'o}tegui, C.}, {\sc Sellmann, M.}, {\sc and} {\sc Tierney, K.} 2009.
\newblock A gender-based genetic algorithm for the automatic configuration of
  algorithms.
\newblock In {\em Proceedings of the Fifteenth International Conference on
  Principles and Practice of Constraint Programming (CP'09)}, {I.~Gent}, Ed.
  Lecture Notes in Computer Science, vol. 5732. Springer-Verlag, 142--157.

\bibitem[\protect\citeauthoryear{Baral}{Baral}{2003}]{baral02a}
{\sc Baral, C.} 2003.
\newblock {\em Knowledge Representation, Reasoning and Declarative Problem
  Solving}.
\newblock Cambridge University Press.

\bibitem[\protect\citeauthoryear{Biere, Heule, {van Maaren}, and Walsh}{Biere
  et~al\mbox{.}}{2009}]{SATHandbook}
{\sc Biere, A.}, {\sc Heule, M.}, {\sc {van Maaren}, H.}, {\sc and} {\sc Walsh,
  T.}, Eds. 2009.
\newblock {\em Handbook of Satisfiability}. Frontiers in Artificial
  Intelligence and Applications, vol. 185.
\newblock IOS Press.

\bibitem[\protect\citeauthoryear{Calimeri, Ianni, Ricca, Alviano, Bria,
  Catalano, Cozza, Faber, Febbraro, Leone, Manna, Martello, Panetta, Perri,
  Reale, Santoro, Sirianni, Terracina, and Veltri}{Calimeri
  et~al\mbox{.}}{2011}]{contest11a}
{\sc Calimeri, F.}, {\sc Ianni, G.}, {\sc Ricca, F.}, {\sc Alviano, M.}, {\sc
  Bria, A.}, {\sc Catalano, G.}, {\sc Cozza, S.}, {\sc Faber, W.}, {\sc
  Febbraro, O.}, {\sc Leone, N.}, {\sc Manna, M.}, {\sc Martello, A.}, {\sc
  Panetta, C.}, {\sc Perri, S.}, {\sc Reale, K.}, {\sc Santoro, M.}, {\sc
  Sirianni, M.}, {\sc Terracina, G.}, {\sc and} {\sc Veltri, P.} 2011.
\newblock The third answer set programming competition: Preliminary report of
  the system competition track.
\newblock See \citeN{lpnmr11}, 388--403.

\bibitem[\protect\citeauthoryear{Coelho, Studer, and Wooldridge}{Coelho
  et~al\mbox{.}}{2010}]{ecai10}
{\sc Coelho, H.}, {\sc Studer, R.}, {\sc and} {\sc Wooldridge, M.}, Eds. 2010.
\newblock {\em Proceedings of the Nineteenth European Conference on Artificial
  Intelligence (ECAI'10)}. IOS Press.

\bibitem[\protect\citeauthoryear{Delgrande and Faber}{Delgrande and
  Faber}{2011}]{lpnmr11}
{\sc Delgrande, J.} {\sc and} {\sc Faber, W.}, Eds. 2011.
\newblock {\em Proceedings of the Eleventh International Conference on Logic
  Programming and Nonmonotonic Reasoning (LPNMR'11)}. Lecture Notes in
  Artificial Intelligence, vol. 6645. Springer-Verlag.

\bibitem[\protect\citeauthoryear{Gagliolo and Schmidhuber}{Gagliolo and
  Schmidhuber}{2006}]{gagsch06a}
{\sc Gagliolo, M.} {\sc and} {\sc Schmidhuber, J.} 2006.
\newblock Learning dynamic algorithm portfolios.
\newblock {\em Annals of Mathematics and Artificial Intelligence\/}~{\em
  47,\/}~3-4, 295--328.

\bibitem[\protect\citeauthoryear{Gebruers, Guerri, Hnich, and Milano}{Gebruers
  et~al\mbox{.}}{2004}]{geguhnmi04a}
{\sc Gebruers, C.}, {\sc Guerri, A.}, {\sc Hnich, B.}, {\sc and} {\sc Milano,
  M.} 2004.
\newblock Making choices using structure at the instance level within a case
  based reasoning framework.
\newblock In {\em Proceedings of the First Conference on Integration of AI and
  OR Techniques in Constraint Programming for Combinatorial Optimization
  Problems (CPAIOR'04)}, {J.~R{\'e}gin} {and} {M.~Rueher}, Eds. Lecture Notes
  in Computer Science, vol. 3011. Springer-Verlag, 380--386.

\bibitem[\protect\citeauthoryear{Gebser, Kaminski, Kaufmann, Ostrowski, Schaub,
  and Schneider}{Gebser et~al\mbox{.}}{2011}]{gekakaosscsc11a}
{\sc Gebser, M.}, {\sc Kaminski, R.}, {\sc Kaufmann, B.}, {\sc Ostrowski, M.},
  {\sc Schaub, T.}, {\sc and} {\sc Schneider, M.} 2011.
\newblock Potassco: The {P}otsdam answer set solving collection.
\newblock {\em AI Communications\/}~{\em 24,\/}~2, 107--124.

\bibitem[\protect\citeauthoryear{Gebser, Kaminski, Kaufmann, Ostrowski, Schaub,
  and Thiele}{Gebser et~al\mbox{.}}{}]{potasscoManual}
{\sc Gebser, M.}, {\sc Kaminski, R.}, {\sc Kaufmann, B.}, {\sc Ostrowski, M.},
  {\sc Schaub, T.}, {\sc and} {\sc Thiele, S.}
\newblock A user's guide to \texttt{gringo}, \texttt{clasp}, \texttt{clingo},
  and \texttt{iclingo}.

\bibitem[\protect\citeauthoryear{Gebser, Kaminski, Kaufmann, and Schaub}{Gebser
  et~al\mbox{.}}{2012}]{gekakasc12a}
{\sc Gebser, M.}, {\sc Kaminski, R.}, {\sc Kaufmann, B.}, {\sc and} {\sc
  Schaub, T.} 2012.
\newblock {\em Answer Set Solving in Practice}.
\newblock Synthesis Lectures on Artificial Intelligence and Machine Learning.
  Morgan and Claypool Publishers.

\bibitem[\protect\citeauthoryear{Gebser, Kaminski, Kaufmann, Schaub, Schneider,
  and Ziller}{Gebser et~al\mbox{.}}{2011}]{gekakascsczi11a}
{\sc Gebser, M.}, {\sc Kaminski, R.}, {\sc Kaufmann, B.}, {\sc Schaub, T.},
  {\sc Schneider, M.}, {\sc and} {\sc Ziller, S.} 2011.
\newblock A portfolio solver for answer set programming: Preliminary report.
\newblock See \citeN{lpnmr11}, 352--357.

\bibitem[\protect\citeauthoryear{Gebser, Kaufmann, and Schaub}{Gebser
  et~al\mbox{.}}{2012}]{gekasc09c}
{\sc Gebser, M.}, {\sc Kaufmann, B.}, {\sc and} {\sc Schaub, T.} 2012.
\newblock Conflict-driven answer set solving: From theory to practice.
\newblock {\em Artificial Intelligence\/}~{\em 187-188}, 52--89.

\bibitem[\protect\citeauthoryear{Gent, Jefferson, Kotthoff, Miguel, Moore,
  Nightingale, and Petrie}{Gent et~al\mbox{.}}{2010}]{gejekomimonipe10a}
{\sc Gent, I.}, {\sc Jefferson, C.}, {\sc Kotthoff, L.}, {\sc Miguel, I.}, {\sc
  Moore, N.}, {\sc Nightingale, P.}, {\sc and} {\sc Petrie, K.} 2010.
\newblock Learning when to use lazy learning in constraint solving.
\newblock See \citeN{ecai10}, 873--878.

\bibitem[\protect\citeauthoryear{Gomes and Selman}{Gomes and
  Selman}{2001}]{gomsel01a}
{\sc Gomes, C.} {\sc and} {\sc Selman, B.} 2001.
\newblock Algorithm portfolios.
\newblock {\em Artificial Intelligence\/}~{\em 126,\/}~1-2, 43--62.

\bibitem[\protect\citeauthoryear{Guerri and Milano}{Guerri and
  Milano}{}]{guemil04a}
{\sc Guerri, A.} {\sc and} {\sc Milano, M.}
\newblock Learning techniques for automatic algorithm portfolio selection.
\newblock 475--479.

\bibitem[\protect\citeauthoryear{Guo and Hsu}{Guo and Hsu}{2004}]{guohsu04}
{\sc Guo, H.} {\sc and} {\sc Hsu, W.} 2004.
\newblock A learning-based algorithm selection meta-reasoner for the real-time
  {MPE} problem.
\newblock In {\em Proceedings of the Seventeenth Australian Joint Conference on
  Artificial Intelligence}. Springer, 307--318.

\bibitem[\protect\citeauthoryear{Hamadi and Schoenauer}{Hamadi and
  Schoenauer}{2012}]{lion12}
{\sc Hamadi, Y.} {\sc and} {\sc Schoenauer, M.}, Eds. 2012.
\newblock {\em Proceedings of the Sixth International Conference Learning and
  Intelligent Optimization (LION'12)}. Lecture Notes in Computer Science, vol.
  7219. Springer-Verlag.

\bibitem[\protect\citeauthoryear{Helmert, R{\"o}ger, and Karpas}{Helmert
  et~al\mbox{.}}{2011}]{heroka11a}
{\sc Helmert, M.}, {\sc R{\"o}ger, G.}, {\sc and} {\sc Karpas, E.} 2011.
\newblock Fast downward stone soup: A baseline for building planner portfolios.
\newblock In {\em ICAPS 2011 Workshop on Planning and Learning}. 28--35.

\bibitem[\protect\citeauthoryear{Hoos, Kaufmann, Schaub, and Schneider}{Hoos
  et~al\mbox{.}}{2013}]{hokascsc13a}
{\sc Hoos, H.}, {\sc Kaufmann, B.}, {\sc Schaub, T.}, {\sc and} {\sc Schneider,
  M.} 2013.
\newblock Robust benchmark set selection for boolean constraint solvers.
\newblock In {\em Proceedings of the Seventh International Conference on
  Learning and Intelligent Optimization (LION'13)}, {P.~Pardalos} {and}
  {G.~Nicosia}, Eds. Lecture Notes in Computer Science. Springer-Verlag, 138--152.

\bibitem[\protect\citeauthoryear{Huberman, Lukose, and Hogg}{Huberman
  et~al\mbox{.}}{1997}]{huluho97a}
{\sc Huberman, B.}, {\sc Lukose, R.}, {\sc and} {\sc Hogg, T.} 1997.
\newblock An economic approach to hard computational problems.
\newblock {\em Science\/}~{\em 275}, 51--54.

\bibitem[\protect\citeauthoryear{Hutter, Hoos, and Leyton-Brown}{Hutter
  et~al\mbox{.}}{2011}]{huhole11b}
{\sc Hutter, F.}, {\sc Hoos, H.}, {\sc and} {\sc Leyton-Brown, K.} 2011.
\newblock Sequential model-based optimization for general algorithm
  configuration.
\newblock In {\em Proceedings of the Fifth International Conference on Learning
  and Intelligent Optimization (LION'11)}. Lecture Notes in Computer Science,
  vol. 6683. Springer-Verlag, 507--523.

\bibitem[\protect\citeauthoryear{Hutter, Hoos, Leyton-Brown, and
  St{\"u}tzle}{Hutter et~al\mbox{.}}{2009}]{huholest09a}
{\sc Hutter, F.}, {\sc Hoos, H.}, {\sc Leyton-Brown, K.}, {\sc and} {\sc
  St{\"u}tzle, T.} 2009.
\newblock Param{ILS}: An automatic algorithm configuration framework.
\newblock {\em Journal of Artificial Intelligence Research\/}~{\em 36},
  267--306.

\bibitem[\protect\citeauthoryear{Hutter, Hoos, and St{\"u}tzle}{Hutter
  et~al\mbox{.}}{2007}]{huhost07a}
{\sc Hutter, F.}, {\sc Hoos, H.}, {\sc and} {\sc St{\"u}tzle, T.} 2007.
\newblock Automatic algorithm configuration based on local search.
\newblock 2007.
\newblock {\em Proceedings of the Twenty-second National Conference on
  Artificial Intelligence (AAAI'07)}. AAAI Press., 1152--1157.

\bibitem[\protect\citeauthoryear{Kadioglu, Malitsky, Sabharwal, Samulowitz, and
  Sellmann}{Kadioglu et~al\mbox{.}}{2011}]{kamasasase11a}
{\sc Kadioglu, S.}, {\sc Malitsky, Y.}, {\sc Sabharwal, A.}, {\sc Samulowitz,
  H.}, {\sc and} {\sc Sellmann, M.} 2011.
\newblock Algorithm selection and scheduling.
\newblock In {\em Proceedings of the Seventeenth International Conference on
  Principles and Practice of Constraint Programming (CP'11)}, {J.~Lee}, Ed.
  Lecture Notes in Computer Science, vol. 6876. Springer-Verlag, 454--469.

\bibitem[\protect\citeauthoryear{Kadioglu, Malitsky, Sellmann, and
  Tierney}{Kadioglu et~al\mbox{.}}{2010}]{kamaseti10a}
{\sc Kadioglu, S.}, {\sc Malitsky, Y.}, {\sc Sellmann, M.}, {\sc and} {\sc
  Tierney, K.} 2010.
\newblock {ISAC} -- instance-specific algorithm configuration.
\newblock See \citeN{ecai10}, 751--756.

\bibitem[\protect\citeauthoryear{L{\'o}pez-Ib{\'a}{\~n}ez, Dubois-Lacoste,
  St{\"u}tzle, and Birattari}{L{\'o}pez-Ib{\'a}{\~n}ez
  et~al\mbox{.}}{2011}]{lodustbi11a}
{\sc L{\'o}pez-Ib{\'a}{\~n}ez, M.}, {\sc Dubois-Lacoste, J.}, {\sc St{\"u}tzle,
  T.}, {\sc and} {\sc Birattari, M.} 2011.
\newblock The irace package, iterated race for automatic algorithm
  configuration.
\newblock Tech. rep., IRIDIA, Universit{\'e} Libre de Bruxelles, Belgium.

\bibitem[\protect\citeauthoryear{Malitsky, Mehta, and O’Sullivan}{Malitsky
  et~al\mbox{.}}{2013}]{mamesu13a}
{\sc Malitsky, Y.}, {\sc Mehta, D.}, {\sc and} {\sc O’Sullivan, B.} 2013.
\newblock Evolving instance specific algorithm configuration.
\newblock In {\em Proceedings of the Sixth Annual Symposium on Combinatorial
  Search (SOCS'13)}, {H.~Helmert} {and} {G.~R{\"o}ger}, Eds. Proceedings of the
  National Conference on Artificial Intelligence (AAAI), 132--140.

\bibitem[\protect\citeauthoryear{Malitsky, Sabharwal, Samulowitz, and
  Sellmann}{Malitsky et~al\mbox{.}}{2012}]{masasase12a}
{\sc Malitsky, Y.}, {\sc Sabharwal, A.}, {\sc Samulowitz, H.}, {\sc and} {\sc
  Sellmann, M.} 2012.
\newblock Parallel sat solver selection and scheduling.
\newblock In {\em Proceedings of the Eighteenth International Conference on
  Principles and Practice of Constraint Programming (CP'12)}, {M.~Milano}, Ed.
  Lecture Notes in Computer Science, vol. 7514. Springer-Verlag, 512--526.

\bibitem[\protect\citeauthoryear{O'Mahony, Hebrard, Holland, Nugent, and
  O'Sullivan}{O'Mahony et~al\mbox{.}}{2008}]{mahehonusu08a}
{\sc O'Mahony, E.}, {\sc Hebrard, E.}, {\sc Holland, A.}, {\sc Nugent, C.},
  {\sc and} {\sc O'Sullivan, B.} 2008.
\newblock Using case-based reasoning in an algorithm portfolio for constraint
  solving.
\newblock In {\em Proceedings of the Nineteenth Irish Conference on Artificial
  Intelligence and Cognitive Science (AICS'08)}, {D.~Bridge}, {K.~Brown},
  {B.~O'Sullivan}, {and} {H.~Sorensen}, Eds.

\bibitem[\protect\citeauthoryear{Pulina and Tacchella}{Pulina and
  Tacchella}{2009}]{pultac09a}
{\sc Pulina, L.} {\sc and} {\sc Tacchella, A.} 2009.
\newblock A self-adaptive multi-engine solver for quantified {B}oolean
  formulas.
\newblock {\em Constraints\/}~{\em 14,\/}~1, 80--116.

\bibitem[\protect\citeauthoryear{Rice}{Rice}{1976}]{rice76a}
{\sc Rice, J.} 1976.
\newblock The algorithm selection problem.
\newblock {\em Advances in Computers\/}~{\em 15}, 65--118.

\bibitem[\protect\citeauthoryear{Roussel}{Roussel}{2011}]{roussel11a}
{\sc Roussel, O.} 2011.
\newblock Description of ppfolio.

\bibitem[\protect\citeauthoryear{Schneider and Hoos}{Schneider and
  Hoos}{2012}]{hoosch11a}
{\sc Schneider, M.} {\sc and} {\sc Hoos, H.} 2012.
\newblock Quantifying homogeneity of instance sets for algorithm configuration.
\newblock See \citeN{lion12}, 190--204.

\bibitem[\protect\citeauthoryear{Seipp, Braun, Garimort, and Helmert}{Seipp
  et~al\mbox{.}}{2012}]{sebrgahe12a}
{\sc Seipp, J.}, {\sc Braun, M.}, {\sc Garimort, J.}, {\sc and} {\sc Helmert,
  M.} 2012.
\newblock Learning portfolios of automatically tuned planners.
\newblock In {\em Proceedings of the Twenty-Second International Conference on
  Automated Planning and Scheduling (ICAPS'12)}, {L.~McCluskey}, {B.~Williams},
  {J.~R. Silva}, {and} {B.~Bonet}, Eds. AAAI, 368--372.

\bibitem[\protect\citeauthoryear{Streeter, Golovin, and Smith}{Streeter
  et~al\mbox{.}}{2007}]{stgosm07a}
{\sc Streeter, M.}, {\sc Golovin, D.}, {\sc and} {\sc Smith, S.} 2007.
\newblock Combining multiple heuristics online. 2007.
\newblock {\em Proceedings of the Twenty-second National Conference on
  Artificial Intelligence (AAAI'07)}. AAAI Press., 1197--1203.

\bibitem[\protect\citeauthoryear{Tamura, Taga, Kitagawa, and Banbara}{Tamura
  et~al\mbox{.}}{2009}]{tatakiba09a}
{\sc Tamura, N.}, {\sc Taga, A.}, {\sc Kitagawa, S.}, {\sc and} {\sc Banbara,
  M.} 2009.
\newblock Compiling finite linear {CSP} into {SAT}.
\newblock {\em Constraints\/}~{\em 14,\/}~2, 254--272.

\bibitem[\protect\citeauthoryear{Xu, Hoos, and Leyton-Brown}{Xu
  et~al\mbox{.}}{2007}]{xuhole07a}
{\sc Xu, L.}, {\sc Hoos, H.}, {\sc and} {\sc Leyton-Brown, K.} 2007.
\newblock Hierarchical hardness models for {SAT}.
\newblock In {\em Proceedings of the Thirteenth International Conference on
  Principles and Practice of Constraint Programming (CP'07)}, {C.~Bessiere},
  Ed. Lecture Notes in Computer Science, vol. 4741. Springer-Verlag, 696--711.

\bibitem[\protect\citeauthoryear{Xu, Hoos, and Leyton-Brown}{Xu
  et~al\mbox{.}}{2010}]{xuhole10a}
{\sc Xu, L.}, {\sc Hoos, H.}, {\sc and} {\sc Leyton-Brown, K.} 2010.
\newblock Hydra: Automatically configuring algorithms for portfolio-based
  selection.
\newblock In {\em Proceedings of the Twenty-fourth National Conference on
  Artificial Intelligence (AAAI'10)}, {M.~Fox} {and} {D.~Poole}, Eds. AAAI
  Press, 210--216.

\bibitem[\protect\citeauthoryear{Xu, Hutter, Hoos, and Leyton-Brown}{Xu
  et~al\mbox{.}}{2008}]{xuhuhole08a}
{\sc Xu, L.}, {\sc Hutter, F.}, {\sc Hoos, H.}, {\sc and} {\sc Leyton-Brown,
  K.} 2008.
\newblock {SAT}zilla: Portfolio-based algorithm selection for {SAT}.
\newblock {\em Journal of Artificial Intelligence Research\/}~{\em 32},
  565--606.

\bibitem[\protect\citeauthoryear{Xu, Hutter, Hoos, and Leyton-Brown}{Xu
  et~al\mbox{.}}{2012}]{xuhuhole11b}
{\sc Xu, L.}, {\sc Hutter, F.}, {\sc Hoos, H.}, {\sc and} {\sc Leyton-Brown,
  K.} 2012.
\newblock Evaluating component solver contributions to portfolio-based
  algorithm selectors.
\newblock In {\em Proceedings of the Fifteenth International Conference on
  Theory and Applications of Satisfiability Testing (SAT'12)}, {A.~Cimatti}
  {and} {R.~Sebastiani}, Eds. Lecture Notes in Computer Science, vol. 7317.
  Springer-Verlag, 228--241.

\bibitem[\protect\citeauthoryear{Yun and Epstein}{Yun and
  Epstein}{2012}]{yuneps12a}
{\sc Yun, X.} {\sc and} {\sc Epstein, S.} 2012.
\newblock Learning algorithm portfolios for parallel execution.
\newblock See \citeN{lion12}, 323--338.

\end{thebibliography}

\end{document}